\documentclass[a4paper,fleqn]{cas-dc}
\usepackage[numbers]{natbib}
\usepackage{subcaption}
\usepackage{pdfcomment}
\usepackage{graphicx}
\usepackage{todonotes}
\usepackage{enumitem}
\usepackage{float}
\def\tsc#1{\csdef{#1}{\textsc{\lowercase{#1}}\xspace}}
\tsc{WGM}
\tsc{QE}

\begin{document}
\let\WriteBookmarks\relax
\def\floatpagepagefraction{1}
\def\textpagefraction{.001}
\let\printorcid\relax 

\shorttitle{Towards Universal Large-Scale Foundational Model for Natural Gas Demand Forecasting}    

\shortauthors{Zhou, Ye, et al.}

\title[mode = title]{Towards Universal Large-Scale Foundational Model for Natural Gas Demand Forecasting}  

\author[1]{Xinxing~Zhou}\fnmark[1]
\author[1]{Jiaqi~Ye}\fnmark[1]
\author[2]{Shubao~Zhao}
\author[3]{Ming~Jin}
\author[2]{Zhaoxiang~Hou}
\author[2]{Chengyi~Yang}
\author[2]{Zengxiang~Li}
\author[1]{Yanlong~Wen} \cormark[1]\ead{wenyl@nankai.edu.cn} 
\author[1]{Xiaojie~Yuan}

\address[1]{College of Computer Science, Nankai University, Tianjin, China} 
\address[2]{Digital Research Insitute, ENN Group, Beijing, China} 
\address[3]{School of Information and Communication Technology, Griffith University, Nathan, Australia}
\fntext[1]{These authors contribute equally.}
\cortext[1]{Corresponding author}  

\begin{abstract}
In the context of global energy strategy, accurate natural gas demand forecasting is crucial for ensuring efficient resource allocation and operational planning. Traditional forecasting methods struggle to cope with the growing complexity and variability of gas consumption patterns across diverse industries and commercial sectors. To address these challenges, we propose the first foundation model specifically tailored for natural gas demand forecasting. Foundation models, known for their ability to generalize across tasks and datasets, offer a robust solution to the limitations of traditional methods, such as the need for separate models for different customer segments and their limited generalization capabilities. Our approach leverages contrastive learning to improve prediction accuracy in real-world scenarios, particularly by tackling issues such as noise in historical consumption data and the potential misclassification of similar data samples, which can lead to degradation in the quaility of the representation and thus the accuracy of downstream forecasting tasks. By integrating advanced noise filtering techniques within the contrastive learning framework, our model enhances the quality of learned representations, leading to more accurate predictions. Furthermore, the model undergoes industry-specific fine-tuning during pretraining, enabling it to better capture the unique characteristics of gas consumption across various sectors. We conducted extensive experiments using a large-scale dataset from ENN Group, which includes data from over 10,000 industrial, commercial, and welfare-related customers across multiple regions. Our model outperformed existing state-of-the-art methods, demonstrating a relative improvement in MSE by 3.68\% and in MASE by 6.15\% compared to the best available model. These results highlight the potential of our approach not only as a powerful tool for demand forecasting in the energy sector but also as a foundation model that can be applied broadly across different contexts, contributing to more reliable and efficient energy management strategies.

\end{abstract}




\maketitle

\section{Introduction}

Natural gas, primarily composed of hydrocarbons, is a crucial energy resource found in underground rock reservoirs. As the cleanest and lowest-carbon fossil fuel, it plays a vital role in reducing environmental pollution and achieving carbon neutrality. With global consumption reaching 3.96 trillion cubic meters in 2023 and continuing to grow in 2024\cite{GJJJ202403003}, the significance of natural gas in the energy sector is undeniable. However, regional imbalances in supply and demand present challenges to resource efficiency and market stability. To address these challenges, accurate demand forecasting is essential. It enables energy companies and policymakers to develop reliable supply plans, manage contracts effectively, and optimize operations, ensuring stable and efficient energy distribution. As technological advancements continue, precise forecasting will become increasingly critical in maximizing the benefits of natural gas.

Many studies \cite{dong2024simmtm, lee2023learning, nie2022time, wu2022timesnet} have focused on improving the accuracy of natural gas demand forecasting, utilizing various advanced methods and algorithms to better capture the complexities of gas consumption.  These approaches have shown effectiveness under controlled conditions, yet they often struggle with real-world industrial applications where additional challenges emerge. Despite the importance of accurate forecasting, two key challenges need to be addressed.  First, industrial datasets often contain significant levels of noise, which can obscure true consumption patterns and complicate accurate demand forecasting.  Noise in the data can stem from various sources, such as sensor errors, data logging issues, or irregular reporting intervals.  These inaccuracies can distort the actual usage patterns, making it difficult to identify genuine consumption trends and anomalies.  Consequently, the presence of noise poses a substantial challenge for modeling and predicting natural gas demand, necessitating robust techniques to filter out these inaccuracies and enhance the reliability of the forecasting model. Second, natural gas consumption patterns vary widely across different sectors, leading to inconsistencies in usage data. In industrial applications, natural gas is used for power generation, heating, steam production, and machinery operation, where usage is often driven by varying production cycles and operational demands, resulting in unpredictable peaks and troughs. In the commercial sector, natural gas is primarily used for cooking, heating, and hot water supply, with demand influenced by business hours and seasonal shifts, such as increased heating needs during colder periods. Residential usage further diversifies these patterns, as natural gas is utilized for household purposes such as fueling water heaters, fireplaces, and stoves, leading to variations in demand based on daily routines and weather conditions. These differences in consumption patterns across industrial, commercial, and residential sectors complicate the modeling process and reduce the predictive accuracy of traditional approaches. Addressing these sector-specific variations is crucial for developing a reliable and accurate demand forecasting model.

The current development of natural gas demand forecasting has evolved from early statistical methods to contemporary deep learning-based approaches. Traditional models, such as time series analysis and regression models, laid the foundation for forecasting by leveraging historical data to identify trends and patterns \cite{erdogdu2010natural, ediger2007arima, aydin2015forecasting, chen2018forecasting}. Recent studies have focused on models such as Recurrent Neural Networks (RNNs) \cite{su2019systematic, laib2019toward}, Convolutional Neural Networks (CNNs) \cite{ding2023forecasting, song2022estimate}, and Transformer \cite{pu2024novel, lin2024compound}, which have shown significant improvements in prediction accuracy. However, most of these approaches still rely on end-to-end learning methods. In recent years, self-supervised representation learning has made significant progress in computer vision (CV) \cite{he2022masked, chen2024context} and natural language processing (NLP) \cite{devlin2019bert, wu2019self}, and it is increasingly being applied to enhance time series prediction as well \cite{nie2022time, lee2023learning}. Given the broad range of natural gas customers and the diversity of their usage patterns, establishing a universal large-scale foundational model for natural gas demand forecasting is essential. Such a model can integrate vast amounts of data from multiple users and regions, enabling more accurate and comprehensive predictions of natural gas consumption. By leveraging advanced machine learning techniques, self-supervised learning techniques and large-scale datasets, our model effectively accounts for the complex dynamics of natural gas demand across diverse geographical and temporal contexts. This foundational model not only enhances the efficiency of natural gas dispatching but also contributes to the overall stability and sustainability of the energy market.

Besides, to address the challenges of noisy industrial datasets and varying consumption patterns across different sectors, we developed this robust foundational model using corresponding  techniques.  First, we introduced a denoising module to filter out noise from the datasets, effectively mitigating the impact of sensor errors, data logging issues, and irregular reporting intervals.  By integrating this denoising module with contrastive learning, we enhanced the model's robustness against noise, ensuring more accurate demand forecasting in industrial contexts.  Second, considering the sector-specific variations in natural gas consumption, we leveraged industry information inherent in our datasets.  We employed a dual criterion of data similarity and industry-specific information to perform negative sample exclusion.  This process helps to avoid misclassifying certain positive samples as negative, thereby preventing the introduction of additional noise into the model.  These techniques collectively enable our model to deliver reliable and precise natural gas demand predictions, accommodating the complex and diverse usage patterns across industrial, commercial, and residential sectors.

The main contributions of this study are as follows:
\begin{enumerate}[label=(\arabic*), itemsep=0pt, parsep=0pt, topsep=0pt]
    \item We have constructed a large-scale, multi-user natural gas demand forecasting dataset, encompassing data from 19 Chinese provinces and over 27 primary industries, containing a total of 13.8M time points. Using this dataset, we developed the first foundational model specifically designed to predict natural gas demand across customers from various cities and industries.
    \item We introduce a denoising module to filter out noise in the data and combine it with contrastive learning to enhance the model's robustness against noise, improving performance on industrial datasets. In contrastive learning, we perform false negative sample removal based on industry information and sample similarity of the dataset to avoid erroneous representation learning and performance degradation due to misclassfication of positive samples as negative ones.
    \item We demonstrate the effectiveness of our model with a 3.68\% improvement in MSE and 6.15\% in MASE, significantly outperforming state-of-the-art methods in natural gas demand forecasting tasks.
\end{enumerate}
\par
The remainder of this paper is organized as follows: Section \ref{Related studies} provides a review of the relevant literature. Section \ref{Methodology} details the methodologies proposed in this study. Section \ref{Experiment} presents the experimental setup, including the dataset, experimental design, evaluation metrics, and results, followed by an analysis of the findings. Finally, Section \ref{Conclusions} offers a summary of the paper's main contributions and conclusions.

\section{Related studies} \label{Related studies}
Methods for forecasting natural gas demand can generally be categorized into statistical methods and AI-based methods \cite{tian2024application}.  Despite significant advancements, challenges remain in applying these methods to large-scale, industrial datasets. This section reviews existing methods and highlights the potential of foundation models in addressing current limitations.
\subsection{Statistical methods}
Statistical methods are widely used for forecasting natural gas demand due to their simplicity and effectiveness.  Common techniques include time series models and regression models. 
Time series models are statistical methods used to analyze and predict future values based on previously observed data points, making them highly applicable in natural gas demand forecasting. When time series data is stationary and free of missing values, the autoregressive integrated moving average (ARIMA) model serves as an effective tool for modeling and predicting data by incorporating autoregressive, differencing, and moving average components \cite{erdogdu2010natural, akpinar2013forecasting, hussain2022time}.  The exponential smoothing method predicts future values by applying weighted averages to historical data, making it suitable for stationary or trend-based time series \cite{akpinar2017day}.  The Vector Autoregression (VAR) model is utilized for multivariate time series analysis, effectively predicting outcomes by capturing the interrelationships between multiple variables \cite{hou2018understanding}.

Regression models are statistical techniques used to predict a dependent variable based on one or more independent variables, commonly applied in natural gas demand forecasting to identify and quantify relationships within the data. Aydin \cite{aydin2015forecasting} found that the S regression model provided the most reliable predictions, validated by statistical approaches. Chen et al.\cite{chen2018forecasting} developed a Functional AutoRegressive model with eXogenous variables (FARX) to provide accurate day-ahead forecasts for the German gas distribution system. Sen et al. \cite{sen2019forecasting} employed various multivariate regression models to forecast future natural gas consumption. 

\subsection{AI-based methods}
Artificial Intelligence (AI) methods have gained popularity in recent years due to their ability to handle complex and non-linear relationships in data. Techniques such as Artificial Neural Networks (ANNs), Support Vector Machines (SVMs), and deep learning approaches have demonstrated promising results in various applications.
Artificial Neural Networks (ANN) are computational models inspired by the human brain, designed to recognize patterns and make predictions by learning from data. Gorucu \cite{gorucu2004artificial} evaluated and forecasted gas consumption in Ankara, Turkey, using Artificial Neural Network (ANN) models.  Khotanzad et al. \cite{khotanzad2000combination} proposed a two-stage system combining two types of artificial neural network (ANN) forecasters to predict daily natural gas consumption. 
Support Vector Machines (SVMs) are supervised learning models used for classification and regression tasks, known for their effectiveness in handling high-dimensional data and nonlinear relationships. Zhu et al. \cite{zhu2015short} introduced a novel approach, the FNF-SVRLP (False Neighbours Filtered-Support Vector Regression Local Predictor), for short-term natural gas demand forecasting. This method integrates SVR with time series reconstruction and optimizes local predictors.  Bai et al. \cite{bai2016daily} propose a structure-calibrated support vector regression (SC-SVR) approach for forecasting daily natural gas consumption, with online calibration of model parameters using an extended Kalman filter.

Building on the success of AI methods, deep learning approaches have emerged as a powerful tool for natural gas demand forecasting, leveraging advanced neural network architectures to capture intricate patterns and dependencies in the data. Merkel et al. \cite{merkel2018short} applied deep neural networks for short-term natural gas load forecasting. Su et al. \cite{su2019systematic} developed a novel Demand Side Management framework for natural gas systems, utilizing a Recurrent Neural Network (RNN) to accurately forecast customer demand based on real-time smart metering data. Laib et al. \cite{laib2019toward} developed a novel hybrid forecasting method combining a Multi-Layered Perceptron (MLP) and Long Short Term Memory (LSTM) models to improve the accuracy of next-day natural gas consumption predictions. Xue et al. \cite{xue2019prediction} proposed an attention gated recurrent unit (AGRU) model for accurately predicting city-level natural gas consumption in district heating systems, leveraging GRU's capability to capture temporal dependencies and reducing the vanishing gradient problem. Petkovic et al. \cite{petkovic2022deep} proposed a deep learning model based on spatio-temporal convolutional neural networks (DLST) to forecast gas flow in Germany's high-pressure transmission network, leveraging CNN's strength in capturing spatial and temporal dependencies. Song et al. \cite{song2022estimate} proposed a hybrid model using seasonal decomposition and temporal convolution network (SDTCN) to predict daily natural gas consumption, leveraging TCN's ability to handle complex, nonlinear time-varying features in large-scale district heating systems. Pu et al. \cite{pu2024novel} developed a spatial-temporal multiscale Transformer network framework combined with a graph neural network (GNN) to enhance the accuracy of short-term natural gas consumption forecasting.  This approach captures dynamic spatial dependencies among users and temporal patterns in historical data, effectively addressing the nonlinear and irregular nature of natural gas consumption. 

In recent years, research on contrastive learning has advanced across various fields\cite{niucontrastive}, with related applications in time series processing\cite{yang2022timeclr} \cite{lee2023learning}. These studies have demonstrated the superiority of contrastive learning in enhancing representation capabilities and its effectiveness in time series analysis, offering a promising new direction for improving the accuracy of natural gas demand forecasting.

\begin{figure*}
    \centering
    \includegraphics[width=0.65\textwidth]{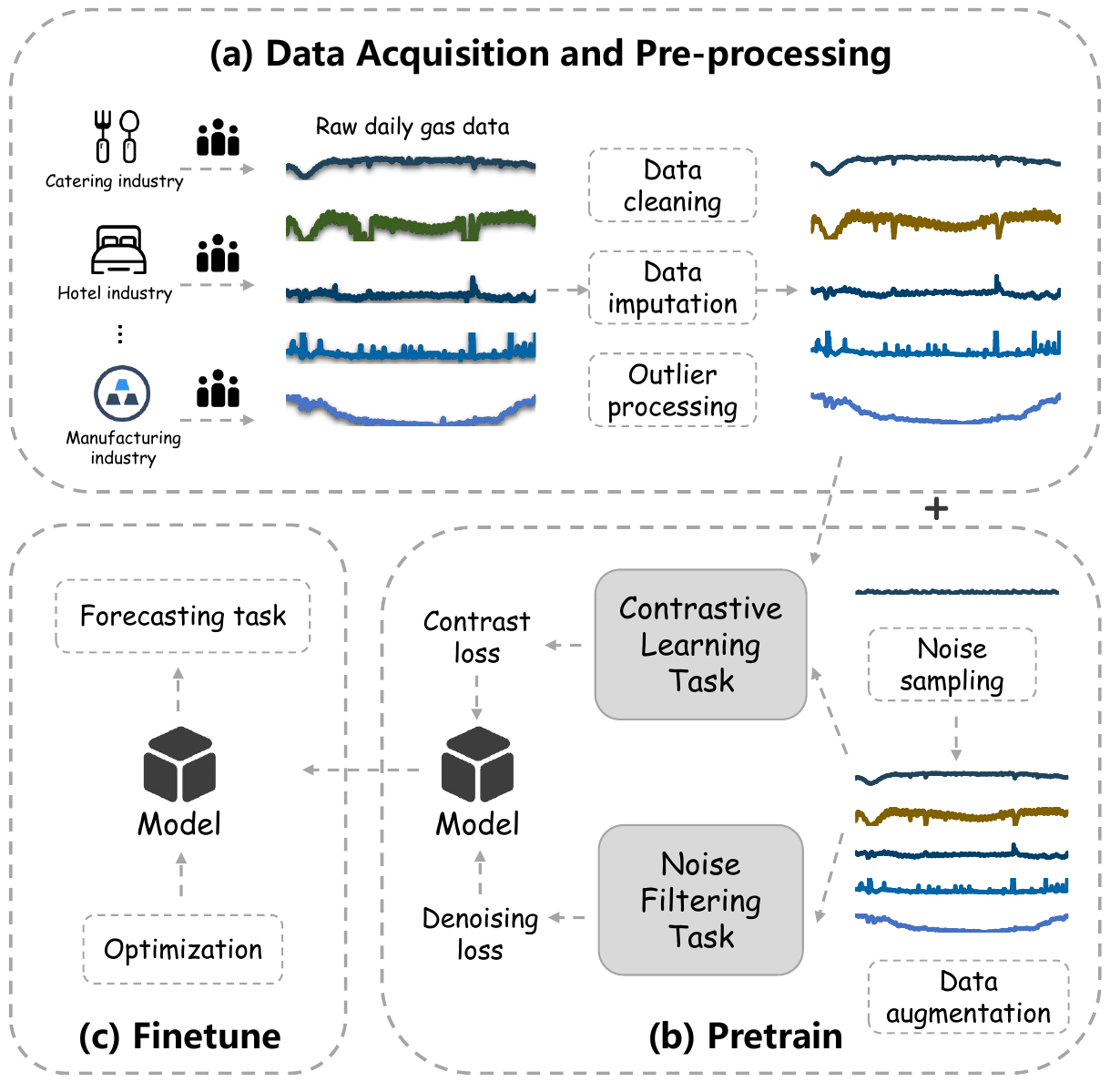}
    \caption{Our workflow, including three phases: (a) data aquisition and pre-processing, (b) pretraining and (c) finetuning for the forecasting task.}
    \label{fig:workflow}
\end{figure*}

\subsection{Foundation Models for Time Series}
Recent advancements in machine learning have introduced the concept of foundation models—large, pre-trained models that can be fine-tuned for specific tasks. Foundation models have demonstrated remarkable generalization capabilities, especially in handling diverse and complex datasets. Darlow et al. \cite{darlow2024dam} proposed the DAM model, a neural network designed for universal forecasting, which excels in zero-shot transfer, long-term forecasting, and handling irregularly sampled data across multiple domains by using randomly sampled histories and a transformer backbone. 
Tu et al. \cite{tu2024powerpm} developed  the PowerPM model for electricity time series (ETS) data. This model is designed with a self-supervised pretraining framework that significantly improves its generalization capabilities across various datasets and applications within the energy sector. Nguyen et al. \cite{nguyen2023climax} introduced ClimaX, a deep learning model designed for weather and climate science. ClimaX leverages the Transformer architecture to handle heterogeneous datasets and is pre-trained on CMIP6 data using self-supervised learning. 

In summary, while various methods have been explored for natural gas demand forecasting—including statistical methods, AI-based techniques, and other innovative approaches—there remain significant challenges in applying these methods to real-world, large-scale datasets. Foundation models, with their ability to generalize across tasks and datasets, offer a powerful solution to these challenges. By addressing the limitations of existing methods, such as their limited generalization capabilities and difficulty handling noisy data, foundation models have the potential to significantly advance the field of natural gas demand forecasting.

\section{Methodology} \label{Methodology}
In this section, we will first introduce our workflow. Then, we will provide detailed sub-points on the design of our data analysis and processing, as well as our pre-training architecture, which includes the industry contrastive learning module and the noise filtering module.
\subsection{Workflow and problem definition}
Figure \ref{fig:workflow} shows our overall workflow. Firstly, we conducted primary data collection from city gas companies across the country. Due to the common problem of missing values, outliers and other problems in industrial datasets, we analyzed and pre-processe the data accordingly to ensure it could be better utilized for training and forecasting tasks. We then perform self-supervised pretraining, a phase that consists of two tasks, contrastive learning and noise filtering, with the aim of enhancing the robustness of the model while giving it better generalization capabilities. Finally, we finetune the model for downstream forecasting task and demonstrate the superiority of our model through its performance on this task.
\par
The time series forecasting task is a classical problem in time series data processing, which involves using the characteristics of one or more time series in a previous period to forecast the characteristics in a future period. A time series of length $t$ can be represented as $X=\{x_1,x_2,...,x_t\}\in \mathbb{R}^{C\times T}$, where $C$ is the number of channels and $T$ is the length of time series. Since our dataset is a univariable multi-user gas dataset, in our dataset $C=1$. We take the historical data of this time series to forecast its future value at $t+1$ step, then result $\hat{X}_{t+1}$ is:
\begin{equation}
    \hat{X}_{t+1}=f(X_{t-n:t})
\end{equation}
where $f(x)$ represents the forecasting model, and$X_{t-n:t}$ represents the time series of $n$ steps before timestamp $t$. 
\subsection{Data acquisition, pre-processing and analysis}
We collected 19,838 records of gas usage data from industrial and commercial users across 84 city gas companies. The data was sourced from various regions nationwide, covering a wide range of industries. Because of the low frequency of gas data generation and the short time span of gas consumption for some industrial and commercial users, some data have the problem of being too short (less than 300 data points) to be used for training and forecasting, we first remove this part of data.
\par
Among the dataset, some users have multiple data entries or data sheets for the same date because they may have multiple gas meters collecting data simultaneously. We consolidated these entries and sheets for each user on the same date and analyzed the data based on the user as a unit rather than the individual gas meter. This approach allows us to leverage user-specific information more efficiently, including industry relevance.
\par
As for outliers, we only removed data with an excessive proportion of positions with too large Z-scores, which are likely to be anomalies due to meter malfunction or transmission delays. The reason that other anomalies were not handled is that the anomalies obtained in the anomaly detection of the time series data are likely to be a true value rather than an anomaly in this dataset, for example, sudden peaks in gas usage or troughs in gas usage, and it is difficult to troubleshoot whether these locations are true or anomalous values. Although these locations may be difficult to forecast, we uniformly leave them unaltered for the sake of data fidelity, which is fair to all baselines. The formula for the Z-score is as follows:
\begin{equation}
    Z=\frac{(x-\mu)}{\sigma}
\end{equation}
where $x$ is the value of the data point, $\mu$ is the mean value of the dataset and $\sigma$ is the standard deviation of the dataset.
\par
\begin{table}
    \centering
    \caption{Details of ENN dataset.}
    \label{tab:details}
    \begin{tabular}{cc}
        \toprule
        Data & Value \\
        \midrule
        Customers & 14817 \\
        Provinces & 19 \\
        Primary Industries & Over 27 \\
        Secondary Industries & Over 120 \\
        Frequency & Daily \\
        Max Series Length & 2355 \\
        Mean Series Length & 931 \\
        Min Series Length & 300 \\
        Time Points & 13.8M \\
        File Size & 59.3M \\
        ADF & -4.51 \\
        \bottomrule
    \end{tabular}
\end{table}
\begin{figure}
    \centering
    \includegraphics[width=0.45\textwidth]{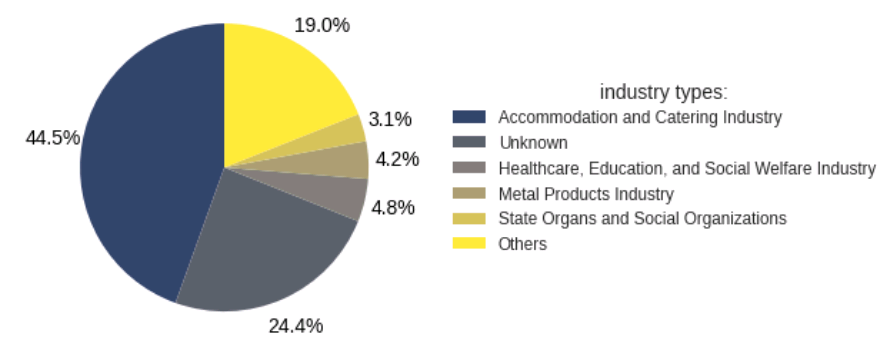}
    \caption{The distribution of users' industries.}
    \label{fig:industry}
\end{figure}
Another problem is missing data, which is a pervasive problem in time series especially in industry datasets. Methods for solving this problem fall into two categories: deletion and imputation. Deleting data, however, leads to incomplete datasets which may result in biased parameter estimates\cite{graham2009missing}. Thus, imputation becomes a better choice. But how and what numbers to fill is a major challenge, while correct imputation estimates are unbiased \cite{white2010bias} and incorrect imputation estimates will introduce extra noise. Many previous works \cite{yu2016temporal} \cite{azur2011multiple} have used statistical or machine learning methods for imputation, but most of them require strong assumptions about the data\cite{cao2018brits}, such as normal distribution or linear regression. These assumptions can lead to biased estimates and affect subsequent analysis results. Recently, much work has been done using deep learning for imputation, achieving state-of-the-art results, so we chose SAITS \cite{du2023saits} as our imputation model. SAITS is a model based only on attention mechanism, which allows it to run faster and be more suitable for our large-scale mutiuser time series data compared to models based on Transformer and Diffusion, and has comparable effects. Since this was not our main work, we simply called the SAITS model for imputation.
\begin{figure}
    \centering
    \includegraphics[width=0.45\textwidth]{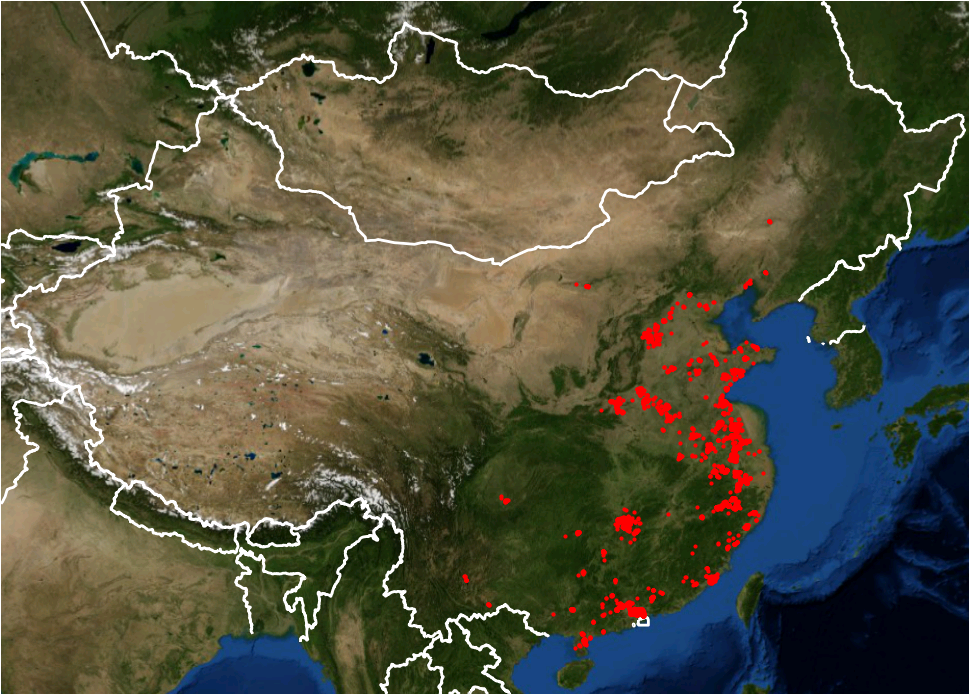}
    \caption{User distribution plotted by the basemap library.}
    \label{fig:map}
\end{figure}
\par
After all these processes, the ENN dataset includes real gas consumption data from 14817 customers. Figure \ref{fig:map} shows the user distribution on the map, which was drawn using the basemap library. And Figure \ref{fig:industry} illustrates the distribution of users' industries. There are a total of more than 27 primary industries and more than 120 sencondary industries, with the accommodation and catering industry accounting for the largest share at 44.5\%, while 24.4\% of the users' industries are unknown. This dataset records daily gas usage data for customers from 2017 to 2024, with records starting and ending at different times for each customer, ranging from a minimum of 300 data entires to a maximum of 2355 data entires, with an average of 931 dataentries. In this dataset, there are a total of 13.8M time points, which is a large dataset in the natural gas field. We stored the date and numerical parts of the dataset in ARROW format, occupying 59.3MB of storage space. To better compare with other dataset, side information such as provinces, cities and industries was not included in the storage calculation. We also computed  the Augmented Dicked-Fuller (ADF) metric to assess the non-stationarity of our dataset. We weighted the ADF metric for each serie according to the data length to obtain an ADF metric of -4.51 for the whole dataset, which proves our dataset is more unsteady and less predictable with higher noise level, so our method can achieve better results.
\subsection{Model architecture}
\begin{figure*}
    \centering
    \includegraphics[width=0.9\textwidth]{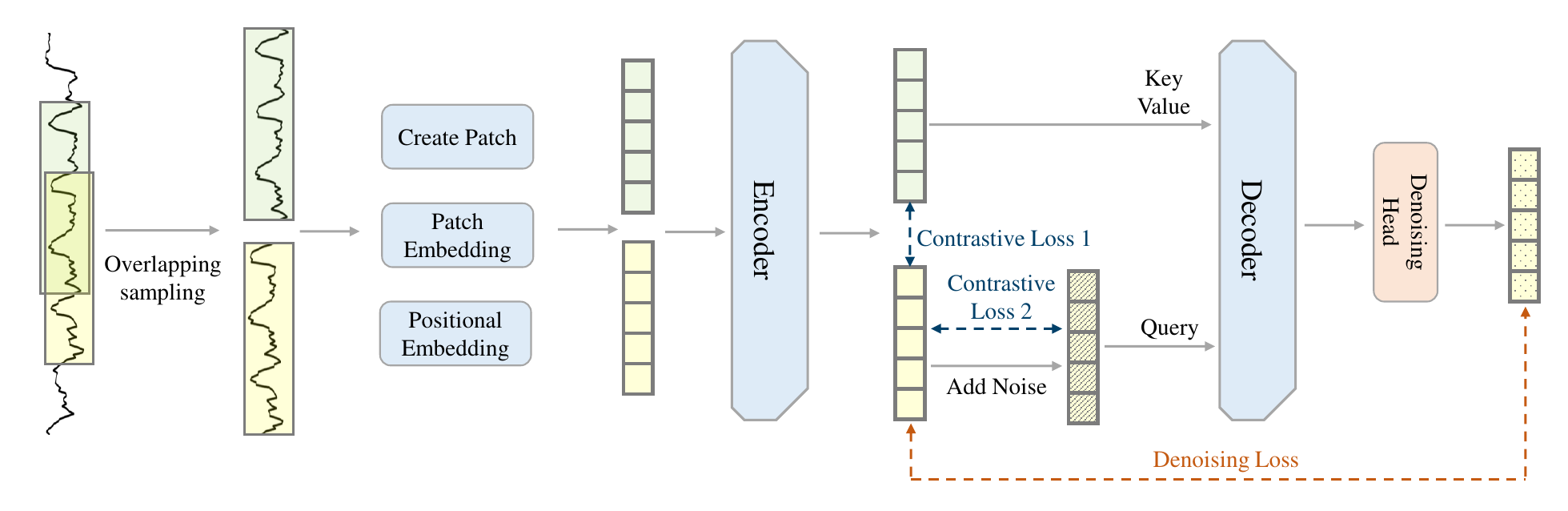}
    \caption{The proposed pre-training model, including patch embedding, data augmentation and loss function. In our Encoder and Decoder, we utilize an attention layer with rotational position encoding from the Llama architecture\cite{touvron2023llama} to achieve better performance. When calculating contrastive loss, only samples from different industries with lower similarity are considered as negative examples.  Samples with higher similarity or from the same industry, which are more likely to result in false negatives, are excluded from both positive and negative categories.  These samples do not participate in the loss calculation, thereby minimizing the introduction of additional noise.}
    \label{fig:model}
\end{figure*}
Figure \ref{fig:model} shows our pre-training model architecture. Our model consists of four components, including patching, contrastive learning, encoder and decoder, which is similar to some contrastive learning methods\cite{lee2023learning} \cite{dong2024simmtm}. However, to the best of our knowledge, previous work has focused on masked modeling, which is helpful to better capture temporal dependencies, but ignores the problem of noise prevalent in industrial datasets. Due to the presence of noise, the model may learn the wrong features and achieve worse results instead. Therefore, we perform data augmentation by adding noise for contrastive learning and design a noise filtering module to enhance the rubustness of the model to noise for better forecasting. In the following, we will introduce each module in the model individuallly.
\par
\subsubsection{Patching.} 
Following the work \cite{nie2022time}, we also use patch embedding because patch embedding helps to preserve local semantic information and can reduce computational complexity and support longer recall windows for better results, and most of the research in recent years has also used patch for embedding\cite{lee2023learning} \cite{zhang2022crossformer} \cite{chen2024pathformer}. Although in many methods, reversible instance normalization (RevIN)\cite{kim2021reversible} was performed before performing patch embedding, but after perform RevIN the metrics declined instead. We believe it is due to the fact that RevIN makes the sequences over smooth, preventing the model from learning useful feautures. Therefore, we directly perform patch embedding and position embedding. 
\begin{equation}
    H_t = Patch\_Embed(X_t) + W_{pos}
\end{equation}
where $H_t$ represents the embedding of a time series data $X_t^i$, and $W_{pos}$ denotes a learnable position embedding, which helps the model recognize the order of elements in a sequence and provide information about the relative positions of data points within the sequence.
\par
\subsubsection{Contrastive learning task with negative sample exclusion} 
Our contrastive learning is divided into two parts, overlapping sampling contrast and noise addition contrast. 
\par
In overlapping sampling contrast, by overlapping samples, we can get more pairs of positive and negative samples and can better utilize the limited data. Contrasting these data can better capture the temporal continuity and enhance the robustness of the model. In a batch, $B$ pairs of overlapping sampled samples can be obtained, with the two sequences in each sample pair denoted as $X_t^i$ and $X_t^{i'}$, which means $2B$ samples in a batch.
\par
Previous work has paid little attention to sequence similarity, often treating all samples other than positive ones as negative samples. This approach overlooks the correlations between sequences caused by similar industry characteristics and gas usage patterns, potentially misleading the representation learning process when construction negative samples \cite{liu2021spatio}. Of course, we cannot directly treat these similar samples as positive samples either, because we cannot ensure that they have feature patterns identical to the original positive samples. Our dataset includes industry information, so we use this information along with sample similarity to guide the selection of negative samples. Specifically, we exclude samples from the same primary industry and the most similar samples as false-negative samples. We first compute the cosine similarity between the samples:
\begin{equation}
    S(H_t^{i}, H_t^{j}) = \frac{H_t^{i}\cdot H_t^{j}}{||H_t^{i}||\ ||H_t^{j}||}
\end{equation}
where $i$ and $j$ represent the $i$th and $j$th samples and $0\leq i,j<2B,\ i\neq j$. False-negative samples can be represented as:
\begin{gather}
    FN_1(i) = \{j|S(H_t^{i}, H_t^{j})\in top(S(H_t^{i}),N)\} \\
    FN_2(i) = \{j|industry(i)==industry(j), j\in N\} \\
    FN(i) = FN_1(i)\cup FN_2(i)
\end{gather}
where N represents the set of negative samples. Then, our overlapped sampling contrastive learning loss can be expressed as follows:
\begin{equation}
    \mathcal{L}_{ssl1} = -\frac{1}{B} \sum_{i=1}^{B} log\frac{exp(H_t^{i}\cdot H_t^{i'})/\tau}{\sum_{j\in 2B\backslash FN(i), j\neq i}exp(H_t^{i}\cdot H_t^{j})/\tau}
\end{equation}
where $B\backslash FN(i)$ means the set of sample in this batch but not in $FN(i)$, and $\tau$ is the temperature parameter.
\par
Inspired by Denoising Diffusion Probabilistic Models (DDPMs) \cite{ho2020denoising}, we introduce a noise filtering task that involves adding noise to the data and learning how to remove it. Unlike DDPMs, which aim for direct forecasting, our method focuses on enhancing the model's robustness to noise. Therefore, it is unnecessary to add noise multiple times until the data becomes pure noise. Instead, we add noise only once, viewing this as a form of data augmentation, which makes it more suitable for integration with contrastive learning. To further strengthen the data representations, we enhance the original data by incorporating a small amount of negative sample embeddings into the positive samples, specifically as outlined below:
\begin{equation}
    (H_t^{i})' = (1-\alpha)H_t^{i} + \alpha H_t^{j}
\end{equation}
where $\alpha$ is a weighting parameter, and $i\neq j$. For convenience, here we did not use negative sample exclusion, which did not significantly affect the results. Then, our noise addition contrastive learning can be expressed as follows:
\begin{equation}
    \mathcal{L}_{ssl2} = -\frac{1}{B} \sum_{i=1}^{B} log\frac{exp(H_t^{i}\cdot (H_t^{i})')/\tau}{\sum_{j=1, j\neq i}^{2B}exp(H_t^{i}\cdot H_t^{j})/\tau}
\end{equation}
\subsubsection{Noise filtering task}
Contrastive learning task can generate better representations, but simply adding noise to contrastive learning does not sufficiently address the challenges posed by noisy data. To further reduce the impact of noise, we design a noise filtering module to counteract this problem. Specifically, we train a decoder and a denoising head to attempt to restore noisy data to its original form, with the denosing head being a multi-layer perceptron (MLP). Due to the strong temporal correlation between the overlapping sampled samples and the noise samples, they can be used to assist in the denoising task. Here are the formulas for self-attention in encoder and decoder: 
\begin{gather}
    Q = (H_t^i)'W^Q,\ K=H_t^{i'}W^K,\ V=H_t^{i'}W^V \\
    Attention(Q,K,V)=softmax(\frac{QK^T}{\sqrt{d_k}})V
\end{gather}
We use Smooth L1 loss to constrain them.
\begin{align}
    \mathcal{L}_{de} = \sum_{i=1}^B
    \begin{cases}
        0.5(\hat{X}_t^i - X_t^i)^2/\beta,\ if|\hat{X}_t^i - X_t^i|<\beta \\
        |\hat{X}_t^i - X_t^i|-0.5\beta,\ if|\hat{X}_t^i - X_t^i|\geq \beta
    \end{cases}
\end{align}
where $\hat{X}_t^i$ is the denoised sequence of the model, and $\beta = 0.01$ is the hyperparameter for Smooth L1 loss.
\subsubsection{Loss function and finetuning}
The overall objective function of our model is a weighted combination of contrastive loss $\mathcal{L}_{ssl1}$, $\mathcal{L}_{ssl2}$ and denoising loss $\mathcal{L}_{de}$ defined as:
\begin{equation}
    \mathcal{L} = (1-\lambda_1-\lambda_2)\mathcal{L}_{de} + \lambda_1 \mathcal{L}_{ssl1} + \lambda_2 \mathcal{L}_{ssl2}
\end{equation}
where $\lambda_1$ and $\lambda_2$ are weighted hyperparameters of the loss. 
\par
In the finetuning phase, we use Mean Squared Error (MSE) loss on the downstream forecasting task for further parameter optimization. Moreover, we use the signal decay-based loss function proposed by research \cite{wang2024card}, as the near-future loss would contribute more to generalization improvement than the far-future loss. Then the loss function becomes as follows:
\begin{equation}
    \mathcal{L} = \frac{1}{B}\sum_{i=1}^B l^{-1/2}(\hat{X}_t^i - X_t^i)^2
\end{equation}
where $l$ is the distance from the starting point of the forecasting. 

\section{Experiment} \label{Experiment}
In this section, we conduct extensive experiments on a dataset collected by several city gas companies to evaluate the effectiveness of the proposed method. We will describe in detail our experimental setup, relative analysis and discussion.
\subsection{Experimental setup}
\subsubsection{Datasets}
Our dataset comes from ENN Energy Holdings Co.Ltd, one of the largest clean energy distributors in China, which has collected a large amount of historical natural gas usage data from consumers in different sectors over the past 30 years. In order to validate the effectiveness of our model on the natural gas dataset, we conducted experiments on nearly 7 years of natural gas consumption  data collected by ENN Energy from various industries.
\par
Due to the characteristics of our dataset, directly following the time series forecasting convention of dividing each customer's data into training, validation and test sets in a 7:1:2 ratio may lead to excessively short validation and test set as well as data leakage problems. Therefore, we use the last six months of data as the test set, and all other data before that are divided into training and validation sets in a 7:1 ratio for a more fair and reasonable experiment.
\subsubsection{Baselines}
We compare our model extensively with 11 recent state-of-art models. Among these, 3 are self-supervised models: PITS\cite{lee2023learning}, SimMTM\cite{dong2024simmtm} and the self-supervised version of PatchTST\cite{nie2022time}. The remaining 9 are end-to-end models: iTransformer\cite{liu2023itransformer}, TimesNet\cite{wu2022timesnet}, DLinear\cite{zeng2023transformers}, Nonstationary Transformer\cite{liu2022non}, Pyraformer\cite{liu2021pyraformer}, FEDformer\cite{zhou2022fedformer}, Autoformer\cite{chen2021autoformer}, the supervised version of PatchTST\cite{nie2022time}, and Informer\cite{zhou2021informer}. 
In the experimental tables of this paper, the best results are highlighted in bold and the second best are underline.

\begin{table*}[htbp]
  \centering
  \caption{Results of medium- and long-term forecasting.  We utilize forecast horizons ranging from 1 to 6 months to better align with realistic application scenarios.  A historical horizon of 96 is selected, consistent with the conventions used in publicly available datasets.}
  \label{tab:long}%
  \resizebox{\linewidth}{!}{
    \begin{tabular}{c|c|cccccccc|cccccc}
    \cmidrule{1-16}
    \multicolumn{2}{c}{\multirow{2}[0]{*}{Models}} & \multicolumn{8}{c|}{Self-supervised}                           & \multicolumn{6}{c}{End-to-End} \\
    \cmidrule{3-16}
    \multicolumn{2}{c}{} & \multicolumn{2}{c}{Ours} & \multicolumn{2}{c}{PITS} & \multicolumn{2}{c}{SimMTM} & \multicolumn{2}{c|}{PatchTST} & \multicolumn{2}{c}{iTransformer} & \multicolumn{2}{c}{PatchTST} & \multicolumn{2}{c}{TimesNet} \\
    \cmidrule{1-16}
    \multicolumn{2}{c}{Metric} & MSE   & MAE   & MSE   & MAE   & MSE   & MAE   & MSE   & MAE   & MSE   & MAE   & MSE   & MAE   & MSE   & MAE \\
    \cmidrule{1-16}
    \multirow{6}[0]{*}{ENN} & 60    & \textbf{0.5001 } & \textbf{0.4888 } & 0.5236  & 0.4967  & 0.5290  & 0.5010  & 0.5234  & 0.4963  & 0.5353  & 0.5049  & 0.5267  & 0.4998  & \underline{0.5163}  & \underline{0.4940}  \\
          & 90    & \textbf{0.5302 } & \textbf{0.5132 } & 0.5846  & 0.5354  & 0.5868  & 0.5379  & 0.5781  & 0.5315  & 0.5607  & 0.5230  & 0.5828  & 0.5357  & 0.5604  & \underline{0.5234}  \\
          & 120   & \textbf{0.5620 } & \textbf{0.5340 } & 0.6398  & 0.5654  & 0.6490  & 0.5710  & 0.6452  & 0.5671  & 0.6096  & 0.5507  & 0.6399  & 0.5670  & 0.6110  & \underline{0.5530}  \\
          & 150   & \textbf{0.5980 } & \textbf{0.5532 } & 0.6894  & 0.5884  & 0.7010  & 0.5950  & 0.6851  & 0.5851  & 0.6623  & 0.5784  & 0.6883  & 0.5896  & 0.6651  & 0.5802  \\
          & 180   & \textbf{0.6343 } & \textbf{0.5702 } & 0.7379  & 0.6080  & 0.7430  & 0.6110  & 0.7415  & 0.6081  & 0.7224  & 0.6036  & 0.7376  & 0.6096  & 0.7457  & 0.6193  \\
    \cmidrule{2-16}
          & Avg   & \textbf{0.5649 } & \textbf{0.5319 } & 0.6351  & 0.5588  & 0.6418  & 0.5632  & 0.6347  & 0.5576  & 0.6181  & 0.5521  & 0.6351  & 0.5603  & 0.6197  & 0.5540  \\
    \cmidrule{1-16}
    \end{tabular}
    }
    \\
    \vspace{1em}
    \resizebox{\textwidth}{!}{
    \begin{tabular}{c|c|cccccccccccccc}
    \cmidrule{1-14}
    \multicolumn{2}{c}{\multirow{2}[0]{*}{Models}} & \multicolumn{12}{c}{End-to-End}                                                               &       &  \\
    \cmidrule{3-14}
    \multicolumn{2}{c}{} & \multicolumn{2}{c}{Dlinear} & \multicolumn{2}{c}{Nonstationary} & \multicolumn{2}{c}{Pyraformer} & \multicolumn{2}{c}{FEDformer} & \multicolumn{2}{c}{Autoformer} & \multicolumn{2}{c}{Informer} &       &  \\
    \cmidrule{1-14}
    \multicolumn{2}{c}{Metric} & MSE   & MAE   & MSE   & MAE   & MSE   & MAE   & MSE   & MAE   & MSE   & MAE   & MSE   & MAE   &       &  \\
    \cmidrule{1-14}
    \multirow{6}[0]{*}{ENN} & 60    & 0.5207  & 0.5102  & 0.5817  & 0.5487  & 0.5201  & 0.5090  & 0.5893  & 0.5532  & 0.7311  & 0.6458  & 0.5662  & 0.5357  &       &  \\
          & 90    & \underline{0.5569}  & 0.5371  & 0.6099  & 0.5697  & 0.5653  & 0.5462  & 0.6299  & 0.5814  & 0.6723  & 0.5981  & 0.6116  & 0.5652  & \phantom{0.0000} & \phantom{0.0000} \\
          & 120   & \underline{0.5877}  & 0.5565  & 0.6661  & 0.6042  & 0.6132  & 0.5782  & 0.6625  & 0.5945  & 0.6933  & 0.6049  & 0.6773  & 0.6088  &       &  \\
          & 150   & \underline{0.6188}  & \underline{0.5726}  & 0.6301  & 0.5800  & 0.6552  & 0.6030  & 0.7066  & 0.6147  & 0.7627  & 0.6416  & 0.6821  & 0.6093  &       &  \\
          & 180   & \underline{0.6485}  & \underline{0.5858}  & 0.6645  & 0.5939  & 0.6960  & 0.6215  & 0.7609  & 0.6338  & 0.8198  & 0.6678  & 0.7155  & 0.6266  &       &  \\
    \cmidrule{2-14}
          & Avg   & \underline{0.5865}  & \underline{0.5524}  & 0.6305  & 0.5793  & 0.6100  & 0.5716  & 0.6698  & 0.5955  & 0.7358  & 0.6316  & 0.6505  & 0.5891  &       &  \\
    \cmidrule{1-14}
    \end{tabular}%
    }
\end{table*}%

\subsubsection{Evaluation metrics}
We use mean square error (MSE) and mean absolute error (MAE), two widely used metrics in time series forecasting, to evaluate the medium- and long-term forecasting capability of the model:
\begin{equation}
    MSE = \frac{1}{n} \sum_{i=1}^{n} (y_i - \hat{y}_i)^2
\end{equation}
\begin{equation}
    MAE = \frac{1}{n} \sum_{i=1}^{n} |y_i - \hat{y}_i|
\end{equation}
The performance of our model over the entire dataset with different forecast lengths and zero-shot capability validation will be evaluated using these two metrics.
\par
For short-term forecasting capability, we will use the commonly employed short-term forecasting metrics SMAPE and MASE for evaluation:
\begin{equation}
    SMAPE = \frac{1}{n} \sum_{i=1}^{n} \frac{|\hat{y}_i-y_i|}{(|\hat{y}_i|+|y_i|)/2}
\end{equation}
\begin{equation}
    MASE = \frac{\frac{1}{n} \sum_{i=1}^{n} |\hat{y}_i-y_i|}{\frac{1}{n-m} \sum_{i=m+1}^{n} |y_i-y_{i-m}|}
\end{equation}
Where $m$ is the seasonal period, because our dataset do not have a clear uniform period, here $m=1$. To better align with real-world applications, users with higher gas consumption should receive more accurate predictions, and the scale of different user data should impact the evaluation metrics. Therefore, our SMAPE calculation includes data denormalization.



\subsection{Results analysis and discussion}

\subsubsection{Main results}

\begin{table*}[htbp]
  \centering
  \caption{Results of short-term forecasting. We selected forecast horizons of one week, half a month, and one month, respectively, while maintaining a historical horizon of 96, consistent with the approach used in other experiments.}
  \label{tab:short}
  \resizebox{\linewidth}{!}{
    \begin{tabular}{c|c|cccccccc|cccccc}
    \cmidrule{1-16}
    \multicolumn{2}{c}{\multirow{2}[0]{*}{Models}} & \multicolumn{8}{c|}{Self-supervised}                           & \multicolumn{6}{c}{End-to-End} \\
    \cmidrule{3-16}
    \multicolumn{2}{c}{} & \multicolumn{2}{c}{Ours} & \multicolumn{2}{c}{PITS} & \multicolumn{2}{c}{SimMTM} & \multicolumn{2}{c|}{PatchTST} & \multicolumn{2}{c}{iTransformer} & \multicolumn{2}{c}{PatchTST} & \multicolumn{2}{c}{TimesNet} \\
    \cmidrule{1-16}
    \multicolumn{2}{c}{Metric} & SMAPE & MASE  & SMAPE & MASE  & SMAPE & MASE  & SMAPE & MASE  & SMAPE & MASE  & SMAPE & MASE  & SMAPE & MASE \\
    \cmidrule{1-16}
    \multirow{4}[0]{*}{ENN} & 7     & \textbf{0.1952}  & \textbf{2.0903}  & 0.1994  & 2.3035  & \underline{0.1964}  & 2.2899  & 0.1972  & 2.2666  & 0.1986  & 2.3074  & 0.1975  & \underline{2.2376}  & 0.1983  & 2.2446  \\
          & 15    & \textbf{0.2035}  & \textbf{1.5106}  & 0.2098  & 1.7327  & \underline{0.2059}  & 1.6719  & 0.2075  & 1.7235  & 0.2075  & 1.7057  & 0.2066  & \underline{1.6510}  & 0.2098  & 1.7468  \\
          & 30    & \textbf{0.2136}  & \textbf{1.4243}  & 0.2203  & 1.5982  & 0.2157  & 1.4859  & 0.2161  & 1.5097  & \underline{0.2144}  & 1.4818  & 0.2162  & \underline{1.4659}  & 0.2167  & 1.4912  \\
    \cmidrule{2-16}
          & Avg   & \textbf{0.2041}  & \textbf{1.6751}  & 0.2098  & 1.8781  & \underline{0.2060}  & 1.8159  & 0.2069  & 1.8333  & 0.2068  & 1.8316  & 0.2068  & \underline{1.7848}  & 0.2083  & 1.8275  \\
    \cmidrule{1-16}
    \end{tabular}
    }
    \\
    \vspace{1em}
  \resizebox{\linewidth}{!}{
    \begin{tabular}{c|c|cccccccccccccc}
    \cmidrule{1-14}
    \multicolumn{2}{c}{\multirow{2}[0]{*}{Models}} & \multicolumn{12}{c}{End-to-End}                                                               &       &  \\
    \cmidrule{3-14}
    \multicolumn{2}{c}{} & \multicolumn{2}{c}{Dlinear} & \multicolumn{2}{c}{Nonstationary} & \multicolumn{2}{c}{Pyraformer} & \multicolumn{2}{c}{FEDformer} & \multicolumn{2}{c}{Autoformer} & \multicolumn{2}{c}{Informer} &       &  \\
    \cmidrule{1-14}
    \multicolumn{2}{c}{Metric} & SMAPE & MASE  & SMAPE & MASE  & SMAPE & MASE  & SMAPE & MASE  & SMAPE & MASE  & SMAPE & MASE  &       &  \\
    \cmidrule{1-14}
    \multirow{4}[0]{*}{ENN} & 7     & 0.2090  & 2.5998  & 0.2077  & 2.4381  & 0.2020  & 2.3225  & 0.2331  & 4.1014  & 0.2309  & 3.7671  & 0.2078  & 2.5128  &       &  \\
          & 15    & 0.2185  & 1.9076  & 0.2195  & 1.8450  & 0.2112  & 1.7722  & 0.2366  & 2.8413  & 0.2443  & 2.8314  & 0.2167  & 1.9172  & \phantom{0.00000} & \phantom{0.00000} \\
          & 30    & 0.2293  & 1.7222  & 0.2293  & 1.6154  & 0.2238  & 1.7832  & 0.2384  & 2.2317  & 0.2539  & 2.3459  & 0.2297  & 1.7988  &       &  \\
    \cmidrule{2-14}
          & Avg   & 0.2189  & 2.0765  & 0.2188  & 1.9662  & 0.2123  & 1.9593  & 0.2360  & 3.0581  & 0.2430  & 2.9815  & 0.2181  & 2.0763  &       &  \\
    \cmidrule{1-14}
    \end{tabular}%
    }
\end{table*}%
The main results of our experiments are divided into long-term and short-term forecasting. Long-term forecasting is crucial for strategic planning and policy-making in the natural gas industry, while short-term forecasting aids in operational efficiency and real-time decision-making. Given the multi-customer nature of our dataset, we evaluate our model's performance in two distinct ways: by comparing the overall mean MSE and MAE against all baselines, and through scatter plots illustrating the performance for each customer against a specific baseline. 

\begin{figure*}[htbp]
  \centering
  \begin{subfigure}[b]{0.45\textwidth}
    \includegraphics[width=\textwidth]{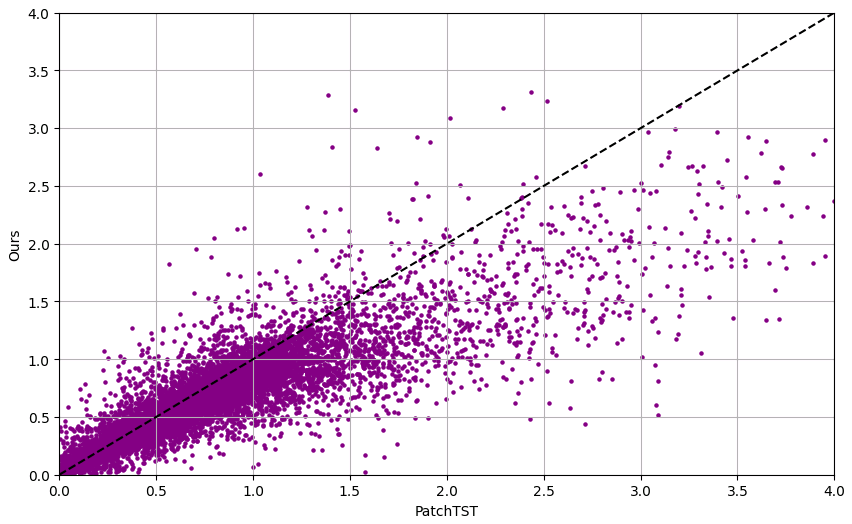}
  \end{subfigure}
  \hfill
  \begin{subfigure}[b]{0.45\textwidth}
    \includegraphics[width=\textwidth]{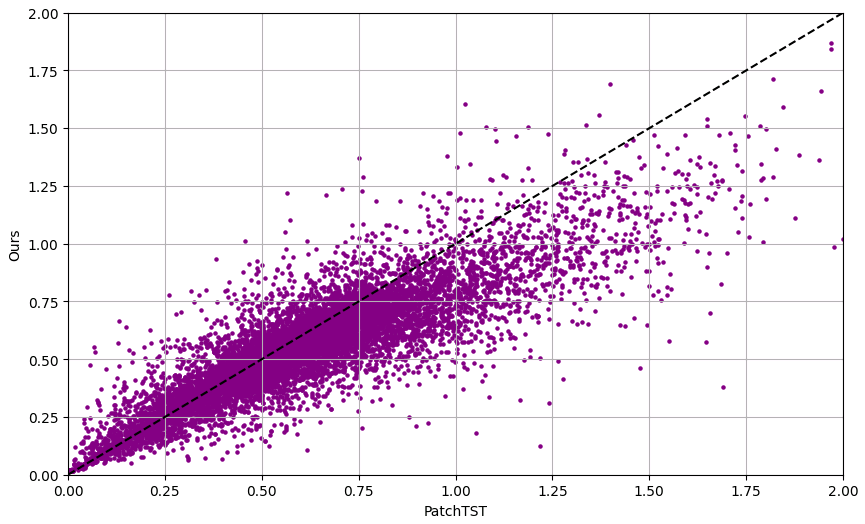}
  \end{subfigure}
  \caption{Model comparison in terms of MSE and MAE.}
  \label{fig:scatter_mse&mae}

  \begin{subfigure}[b]{0.45\textwidth}
    \includegraphics[width=\textwidth]{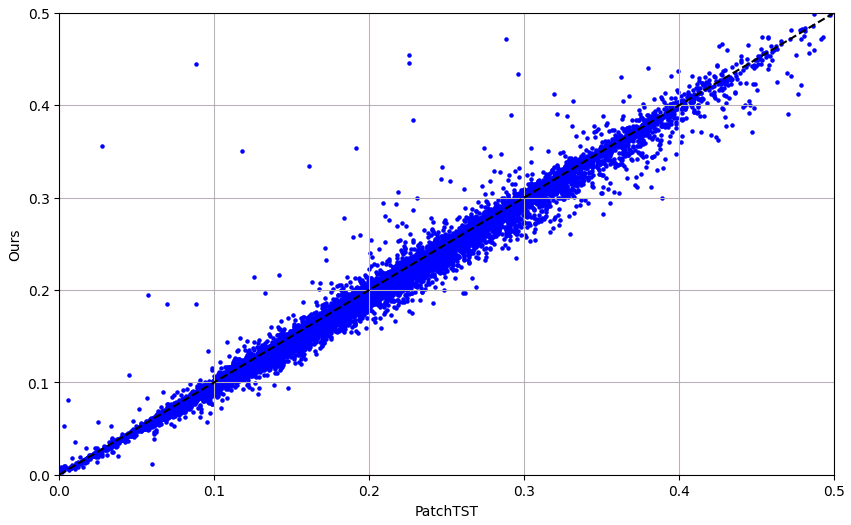}
  \end{subfigure}
  \hfill
  \begin{subfigure}[b]{0.45\textwidth}
    \includegraphics[width=\textwidth]{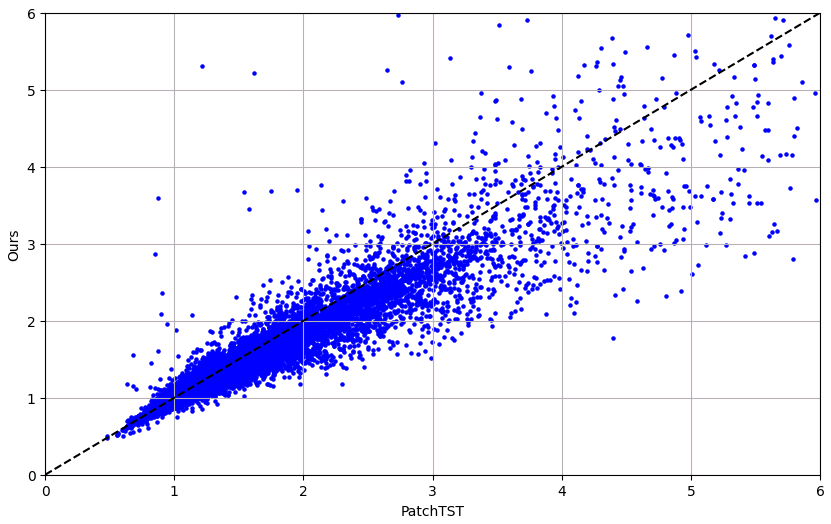}
  \end{subfigure}
  \caption{Model comparison in terms of SMAPE and MASE.}
  \label{fig:scatter_smape&mase}
\end{figure*}

For long-term forecasting, we set the prediction horizons to 60, 90, 120, 150, and 180 days. As shown in Table ~\ref{tab:long}, the accuracy of the model predictions decreases as the prediction horizon increases, while the lower MSE and MAE values indicate better prediction accuracy. Our model is compared against eight end-to-end models and three self-supervised learning baselines. The results demonstrate that our model consistently outperforms all others across all prediction horizons, highlighting its significant advantage in long-term forecasting. Compared to the best-performing baseline model, our model demonstrates a 3.68\% improvement in the MSE metric and a 3.71\% improvement in the MAE metric. Figure \ref{fig:scatter_mse&mae} provides scatter plots that further illustrate the comparative performance of our model against PatchTST. In these plots, each point represents a single customer, with the horizontal axis showing the metrics obtained using the PatchTST model and the vertical axis showing the metrics obtained using our model. The concentration of points below the y=x line indicates that our model achieves lower MSE and MAE values for the majority of customers, demonstrating superior performance. This visual representation confirms that our model consistently outperforms PatchTST, providing more accurate long-term forecasting results across a diverse set of customers.

In our short-term forecasting analysis, we examined prediction horizons of 7, 15, and 30 days. As detailed in Table~\ref{tab:short}, our model maintains high predictive accuracy across these shorter intervals. The comparison with the same eleven baselines reveals that our model consistently delivers superior results. Specifically, the improvements in MASE metrics, relative to the best baseline models, reached 6.15\% . Figure \ref{fig:scatter_smape&mase} presents scatter plots for short-term forecasting. The concentration of points below the y=x line indicates that our model achieves lower SMAPE and MASE values for the majority of customers. This visual representation highlights our model’s ability to deliver more accurate short-term forecasts, which is crucial for optimizing daily operations and facilitating immediate decision-making processes.

\subsubsection{Transfer learning scenarios} 
\begin{table*}[htbp]
  \centering
  \caption{Results of transfer experiments. We pre-trained on one subset, followed by fine-tuning and testing on another. For short-term forecasting, we used prediction steps of 7, 15, and 30, while for medium- to long-term forecasting, prediction steps of 60, 90, and 120 were selected.}
  \label{tab:transfer}
  \resizebox{\linewidth}{!}{
    \begin{tabular}{c|c|cccccccc|ccccccccc}
    \midrule
    \multicolumn{2}{c}{\multirow{2}[0]{*}{Models}} & \multicolumn{8}{c|}{Short-term}                                & \multicolumn{9}{c}{Medium- and long-term} \\
    \cmidrule{3-19}
    \multicolumn{2}{c}{} & \multicolumn{2}{c}{Ours} & \multicolumn{2}{c}{PITS} & \multicolumn{2}{c}{PatchTST} & \multicolumn{2}{c|}{SimMTM} &  & \multicolumn{2}{c}{Ours} & \multicolumn{2}{c}{PITS} & \multicolumn{2}{c}{PatchTST} & \multicolumn{2}{c}{SimMTM} \\
    \midrule
    \multicolumn{2}{c}{Metric} & SMAPE & MASE  & SMAPE & MASE  & SMAPE & MASE  & SMAPE & MASE &  & MSE   & MAE   & MSE   & MAE   & MSE   & MAE   & MSE   & MAE \\
    \midrule
    \multicolumn{1}{c|}{\multirow{4}{*}{\parbox{1cm}{\centering Part I\newline{} $\downarrow$\newline{}Part II}}} & 7     & \textbf{0.1970 } & \textbf{2.1284 } & 0.2020  & 2.4330  & 0.1996  & 2.4184  & 0.1997  & 2.4993  &  \multicolumn{1}{c|}{60} & \textbf{0.5081 } & \textbf{0.4853 } & 0.5187  & 0.5040  & 0.5115  & 0.4983  & 0.5350  & 0.5080  \\
          & 15    & \textbf{0.2080 } & \textbf{1.7042 } & 0.2116  & 1.8097  & 0.2104  & 1.8340  & 0.2088  & 1.8733 &  \multicolumn{1}{c|}{90} & \textbf{0.5397 } & \textbf{0.5084 } & 0.5496  & 0.5272  & 0.5448  & 0.5236  & 0.5860  & 0.5407  \\
          & 30    & \textbf{0.2133 } & \textbf{1.4394 } & 0.2214  & 1.5893  & 0.2201  & 1.5905  & 0.2187  & 1.6254 &  \multicolumn{1}{c|}{120} & \textbf{0.5720 } & \textbf{0.5285 } & 0.5807  & 0.5479  & 0.5750  & 0.5422  & 0.6343  & 0.5676  \\
    \cmidrule{2-19}
          & Avg   & \textbf{0.2061 } & \textbf{1.7573 } & 0.2117  & 1.9440  & 0.2100  & 1.9476  & 0.2091  & 1.9993 &  \multicolumn{1}{c|}{Avg} & \textbf{0.5399 } & \textbf{0.5074 } & 0.5497  & 0.5264  & 0.5438  & 0.5214  & 0.5851  & 0.5388  \\
    \midrule
    \end{tabular}%
    }
    \\
    \vspace{1em}
  \resizebox{\linewidth}{!}{
    \begin{tabular}{c|c|cccccccc|ccccccccc}
    \midrule
    \multicolumn{2}{c}{\multirow{2}[0]{*}{Models}} & \multicolumn{8}{c|}{Short-term}                                & \multicolumn{9}{c}{Medium- and long-term} \\
    \cmidrule{3-19}
    \multicolumn{2}{c}{} & \multicolumn{2}{c}{Ours} & \multicolumn{2}{c}{PITS} & \multicolumn{2}{c}{PatchTST} & \multicolumn{2}{c|}{SimMTM} & & \multicolumn{2}{c}{Ours} & \multicolumn{2}{c}{PITS} & \multicolumn{2}{c}{PatchTST} & \multicolumn{2}{c}{SimMTM} \\
    \midrule
    \multicolumn{2}{c}{Metric} & SMAPE & MASE  & SMAPE & MASE  & SMAPE & MASE  & SMAPE & MASE  & & MSE   & MAE   & MSE   & MAE   & MSE   & MAE   & MSE   & MAE \\
    \midrule
    \multicolumn{1}{c|}{\multirow{4}{*}{\parbox{1cm}{\centering Part II\newline{}$\downarrow$\newline{}Part I}}} & 7     & \textbf{0.1924 } & \textbf{1.9788 } & 0.1997  & 2.3443  & 0.1972  & 2.1942  & 0.1956  & 2.1681  &  \multicolumn{1}{c|}{60} & \textbf{0.4910 } & \textbf{0.4729 } & 0.4976  & 0.4898  & 0.4938  & 0.4858  & 0.5241  & 0.4958  \\
          & 15    & \textbf{0.2017 } & \textbf{1.4482 } & 0.2086  & 1.7034  & 0.2057  & 1.5980  & 0.2047  & 1.5309 &  \multicolumn{1}{c|}{90} & \textbf{0.5215 } & \textbf{0.4962 } & 0.5298  & 0.5154  & 0.5266  & 0.5129  & 0.5850  & 0.5334  \\
          & 30    & \textbf{0.2088 } & \textbf{1.3147 } & 0.2200  & 1.6129  & 0.2158  & 1.4895  & 0.2147  & 1.4046 &  \multicolumn{1}{c|}{120} & \textbf{0.5601 } & \textbf{0.5187 } & 0.5637  & 0.5378  & 0.5610  & 0.5346  & 0.6564  & 0.5715  \\
    \cmidrule{2-19}
          & Avg   & \textbf{0.2010 } & \textbf{1.5806 } & 0.2094  & 1.8869  & 0.2062  & 1.7606  & 0.2050  & 1.7012 &  \multicolumn{1}{c|}{Avg} & \textbf{0.5242 } & \textbf{0.4959 } & 0.5304  & 0.5143  & 0.5271  & 0.5111  & 0.5885  & 0.5336  \\
    \midrule
    \end{tabular}%
    }
\end{table*}%
\label{sec:transfer}
In this section, we explore the performance of our model under transfer learning scenarios, which were designed to assess its robustness and generalizability. The transfer learning experiments involved randomly partitioning the dataset into two distinct subsets, Part I and Part II. Specifically, each customer in the dataset was randomly assigned a binary label (0 or 1), with those labeled as 0 being allocated to Part I and those labeled as 1 to Part II. After partitioning, Part I contained data from 8,139 users, while Part II had data from 6,678 users. This random assignment allowed us to conduct pre-training on one subset before fine-tuning the model on the other, effectively simulating a real-world scenario where a model trained on one group of customers is adapted for use on a different group.

For the transfer learning experiments, we compared our model against three state-of-the-art self-supervised learning baselines: PITS, PatchTST, and SimMTM. We evaluated model performance in both short-term and medium- to long-term forecasting contexts. The short-term forecasting task focused on prediction horizons of 7, 15, and 30 days, while the medium- and long-term forecasting task examined horizons of 60, 90, and 120 days.

Table~\ref{tab:transfer} summarizes the transfer learning results. Across all scenarios, our model consistently outperforms the baseline methods, achieving the best results in both SMAPE and MASE for short-term forecasting, as well as MSE and MAE for medium- and long-term forecasting. This superior performance underscores the effectiveness of our model's architecture, which is particularly well-suited for capturing the complex temporal dependencies and variability inherent in the data.

Our model demonstrates significant performance improvements across both short-term and long-term forecasting horizons when compared to baseline models. For instance, in the 7-day horizon of the short-term predictions, our model achieves up to a 12.52\% reduction in MASE and a 2.48\% reduction in SMAPE compared to PITS. Across all short-term horizons, our model shows an average MASE improvement of 9.77\% over PatchTST and 12.1\% over SimMTM. In the medium- and long-term horizons, our model maintains its competitive edge, with a consistent lead in predictive accuracy. Notably, in the 90-day forecast, our model improves MSE by 0.97\% and MAE by 3.26\% compared to the best baseline, PatchTST. While the numerical improvements in these longer horizons may appear modest, they are critical for applications requiring sustained accuracy over extended periods, reinforcing the model's capability to deliver reliable forecasts even as the prediction window lengthens.

Our comparative analysis demonstrates that our model consistently surpasses baseline methods across all metrics and forecasting horizons.  This capability is particularly valuable in natural gas demand forecasting, where data from different regions or periods may vary significantly. By leveraging pre-trained knowledge, our model can more accurately predict demand in new or less-represented contexts, leading to more efficient resource management and operational planning.

\subsubsection{Zero-shot scenarios} 
\begin{table*}[htbp]
  \centering
  \caption{Results of zero-shot experiments. We pre-trained and fine-tuned on one subset, then tested on another. The forecast and historical horizons used were consistent with those in the transfer experiments.}
  \label{tab:zero}
  \resizebox{\linewidth}{!}{
    \begin{tabular}{c|c|cccccccc|ccccccccc}
    \midrule
    \multicolumn{2}{c}{\multirow{2}[0]{*}{Models}} & \multicolumn{8}{c|}{Short-term}                                & \multicolumn{9}{c}{Medium- and long-term} \\
    \cmidrule{3-19}
    \multicolumn{2}{c}{} & \multicolumn{2}{c}{Ours} & \multicolumn{2}{c}{PITS} & \multicolumn{2}{c}{PatchTST} & \multicolumn{2}{c|}{SimMTM} &  & \multicolumn{2}{c}{Ours} & \multicolumn{2}{c}{PITS} & \multicolumn{2}{c}{PatchTST} & \multicolumn{2}{c}{SimMTM} \\
    \midrule
    \multicolumn{2}{c}{Metric} & SMAPE & MASE  & SMAPE & MASE  & SMAPE & MASE  & SMAPE & MASE &  & MSE   & MAE   & MSE   & MAE   & MSE   & MAE   & MSE   & MAE \\
    \midrule
    \multicolumn{1}{c|}{\multirow{4}{*}{\parbox{1cm}{\centering Part I\newline{} $\downarrow$\newline{}Part II}}} & 7 & \textbf{0.1980 } & \textbf{2.1175 } & 0.2026  & 2.4234  & 0.2003  & 2.3394  & 0.1990  & 2.4360 & \multicolumn{1}{c|}{60} & \textbf{0.5120 } & \textbf{0.4873 } & 0.5218  & 0.5068  & 0.5169  & 0.5025  & 0.5359  & 0.5063  \\
    & 15 & \textbf{0.2072 } & \textbf{1.6364 } & 0.2135  & 1.8657  & 0.2102  & 1.7581  & 0.2088  & 1.8604 & \multicolumn{1}{c|}{90} & \textbf{0.5469 } & \textbf{0.5132 } & 0.5536  & 0.5313  & 0.5516  & 0.5292  & 0.5896  & 0.5401  \\
    & 30 & \textbf{0.2139 } & \textbf{1.4351 } & 0.2241  & 1.6395  & 0.2200  & 1.5737  & 0.2189  & 1.6089  & \multicolumn{1}{c|}{120} & \textbf{0.5821 } & \textbf{0.5340 } & 0.5864  & 0.5533  & 0.5825  & 0.5501  & 0.6464  & 0.5718  \\
    \cmidrule{2-19}
    & Avg & \textbf{0.2064 } & \textbf{1.7297 } & 0.2134  & 1.9762  & 0.2102  & 1.8904  & 0.2089  & 1.9684 & \multicolumn{1}{c|}{Avg} & \textbf{0.5470 } & \textbf{0.5115 } & 0.5539  & 0.5305  & 0.5503  & 0.5273  & 0.5906  & 0.5394  \\
    \midrule
    \end{tabular}%
    }
    \\
    \vspace{1em}
  \resizebox{\linewidth}{!}{
    \begin{tabular}{c|c|cccccccc|ccccccccc}
    \midrule
    \multicolumn{2}{c}{\multirow{2}[0]{*}{Models}} & \multicolumn{8}{c|}{Short-term}                                & \multicolumn{9}{c}{Medium- and long-term} \\
    \cmidrule{3-19}
    \multicolumn{2}{c}{} & \multicolumn{2}{c}{Ours} & \multicolumn{2}{c}{PITS} & \multicolumn{2}{c}{PatchTST} & \multicolumn{2}{c|}{SimMTM} & & \multicolumn{2}{c}{Ours} & \multicolumn{2}{c}{PITS} & \multicolumn{2}{c}{PatchTST} & \multicolumn{2}{c}{SimMTM} \\
    \midrule
    \multicolumn{2}{c}{Metric} & SMAPE & MASE  & SMAPE & MASE  & SMAPE & MASE  & SMAPE & MASE  & & MSE   & MAE   & MSE   & MAE   & MSE   & MAE   & MSE   & MAE \\
    \midrule
    \multicolumn{1}{c|}{\multirow{4}{*}{\parbox{1cm}{\centering Part II\newline{}$\downarrow$\newline{}Part I}}} & 7     & \textbf{0.1949 } & \textbf{2.1130 } & 0.1984  & 2.2797  & 0.1958  & 2.2155  & 0.1965 & 2.2227 & \multicolumn{1}{c|}{60}    & \textbf{0.4906 } & \textbf{0.4726 } & 0.4972  & 0.4883  & 0.4931  & 0.4845  & 0.5263 & 0.4988 \\
    & 15    & \textbf{0.2009 } & \textbf{1.4356 } & 0.2072  & 1.6400  & 0.2057  & 1.6291  & 0.2050 & 1.5867 & \multicolumn{1}{c|}{90}    & \textbf{0.5188 } & \textbf{0.5001 } & 0.5320  & 0.5148  & 0.5235  & 0.5082  & 0.5854 & 0.5358 \\
    & 30    & \textbf{0.2094 } & \textbf{1.3226 } & 0.2177  & 1.5400  & 0.2147  & 1.4798  & 0.2150 & 1.4453 & \multicolumn{1}{c|}{120}   & \textbf{0.5566 } & \textbf{0.5235 } & 0.5649  & 0.5350  & 0.5584  & 0.5289  & 0.6462 & 0.5682 \\
    \cmidrule{2-19}
    & Avg   & \textbf{0.2017 } & \textbf{1.6237 } & 0.2078  & 1.8199  & 0.2054  & 1.7748  & 0.2055 & 1.7516 & \multicolumn{1}{c|}{Avg}   & \textbf{0.5220 } & \textbf{0.4987 } & 0.5314  & 0.5127  & 0.5250  & 0.5072  & 0.5860 & 0.5343 \\

    \midrule
    \end{tabular}%
    }
\end{table*}%
In this section, we explore the performance of our model in zero-shot scenarios, which helps to validate the model's prediction accuracy for unseen user sequences and better use in real application scenarios. Similar to transfer learning scenarios, the zero-shot scenarios also require two subsets for the corresponding experiments, so we use the same randomly partitioned subsets Part I and Part II as in section \ref{sec:transfer}. 
\par
Also in a similar setup to the transfer learning scenario, we compared our models to three state-of-the-art self-supervised learning baselines: PITS, PatchTST, and SimMTM. We pre-trained and fine-tuned all of these models on one subset, and tested them on another subset, which corresponded to the prediction of gas usage of newly added customers in a real scenario. We similarly evaluated model capabilities on short-term prediction tasks with prediction steps of 7, 15, and 30 and on medium- and long-term prediction tasks with prediction steps of 60, 90, and 120.
\par
The performance of our model compared to the baselines is shown in Table \ref{tab:zero}. As shown in the table, our model consistently outperforms the baseline methods on all metrics. This demonstrates that our model effectively generalizes from large volumes of time-series data, enabling it to make more accurate predictions for user data not included in the training set.
\par
Specifically, across all short-term horizons, our model's SMAPE metric outperforms PITS by 3.2\%, PatchTST by 1.9\%, and SimMTM by 1.6\% on average. In terms of the MASE metric, this advantage is even more pronounced at 11.7\%, 9.6\% and 9.7\%. On all medium- and long-term horizons, the MSE and MAE metrics improve less because the models have not seen the training data from the test set users and are more difficult to predict, but there is still a steady improvement over all baselines. For example, compared to PatchTST, which is the best performer in the baseline, MSE and MAE improve by 0.6\% and 2.4\%, respectively. 
\par
It is also worth noting that in almost all the experiments especially in the zero-shot scenario, PatchTST shows better or equivalent performance than PITS and SimMTM, which suggests that the pre-training strategies of PITS and SimMTM do not show significant advantages or even bring disadvantages in the present dataset due to the influence of noisy data and other reasons.
\par
Our experiments and analyses demonstrate that our model outperforms others in zero-short scenarios across all metrics and prediction horizons. These zero-shot experiments not only highlight the strong generalization ability of our model but also carry significant implications for real-world applications. Specifically, our model proves to be more efficient for new users and those with limited historical data. This capability allows for more accurate predictions for such users, leading to better forcasts of gas consumption, which can aid in resource management and scheduling. As a result, our model shows great potential for application.

\begin{figure*}[htbp]
  \centering
  \begin{subfigure}[b]{0.32\textwidth}
    \includegraphics[width=\textwidth]{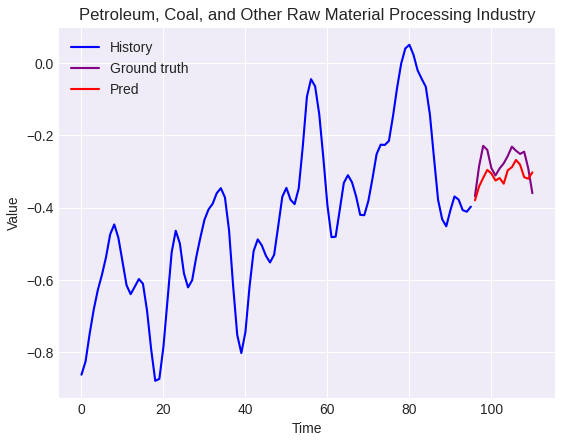}
    \label{fig:subfig1}
  \end{subfigure}
  \hfill
  \begin{subfigure}[b]{0.32\textwidth}
    \includegraphics[width=\textwidth]{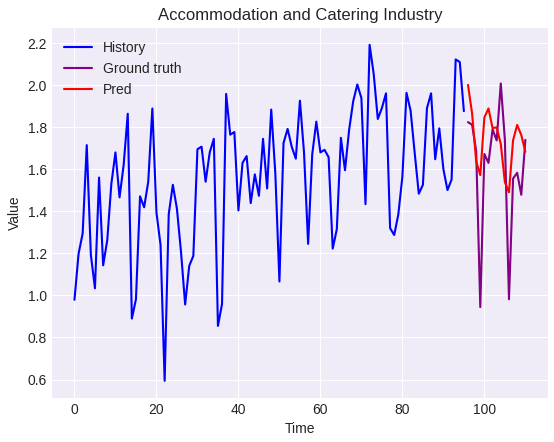}
    \label{fig:subfig2}
  \end{subfigure}
  \hfill
  \begin{subfigure}[b]{0.32\textwidth}
    \includegraphics[width=\textwidth]{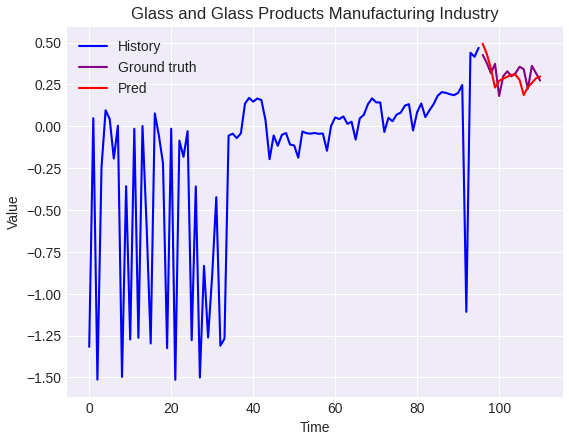}
    \label{fig:subfig3}
  \end{subfigure}
  15-day forecasting of our model.
  \vspace{1em}
  
  \begin{subfigure}[b]{0.32\textwidth}
    \includegraphics[width=\textwidth]{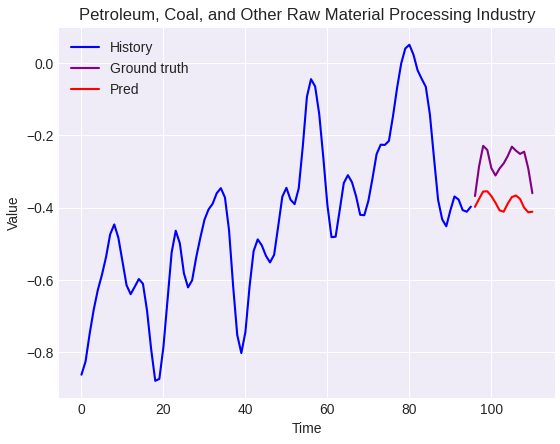}
    \label{fig:subfig4}
  \end{subfigure}
  \hfill
  \begin{subfigure}[b]{0.32\textwidth}
    \includegraphics[width=\textwidth]{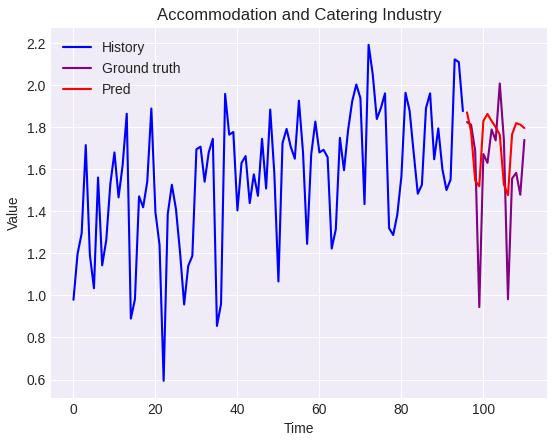}
    \label{fig:subfig5}
  \end{subfigure}
  \hfill
  \begin{subfigure}[b]{0.32\textwidth}
    \includegraphics[width=\textwidth]{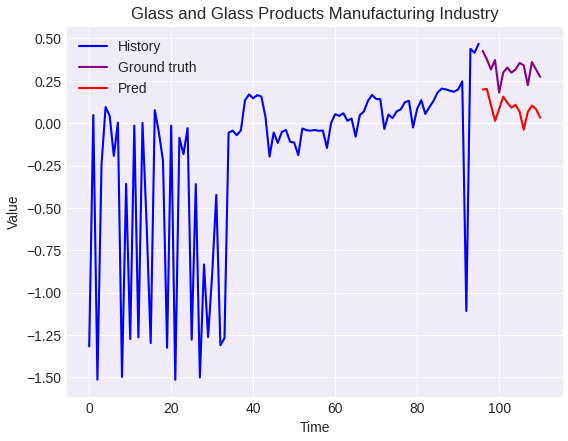}
    \label{fig:subfig6}
  \end{subfigure}
  15-day forecasting of patchtst.
  \vspace{1em}
  
  \begin{subfigure}[b]{0.32\textwidth}
    \includegraphics[width=\textwidth]{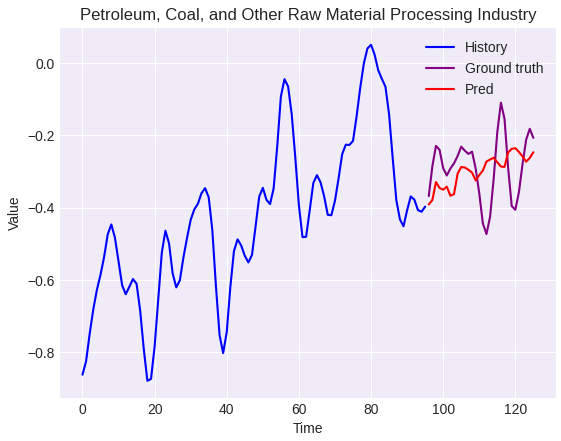}
    \label{fig:subfig7}
  \end{subfigure}
  \hfill
  \begin{subfigure}[b]{0.32\textwidth}
    \includegraphics[width=\textwidth]{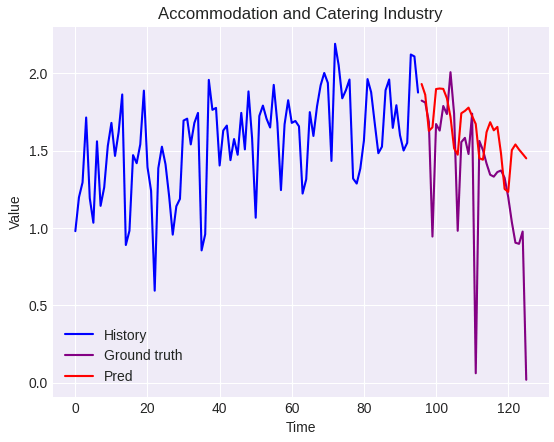}
    \label{fig:subfig8}
  \end{subfigure}
  \hfill
  \begin{subfigure}[b]{0.32\textwidth}
    \includegraphics[width=\textwidth]{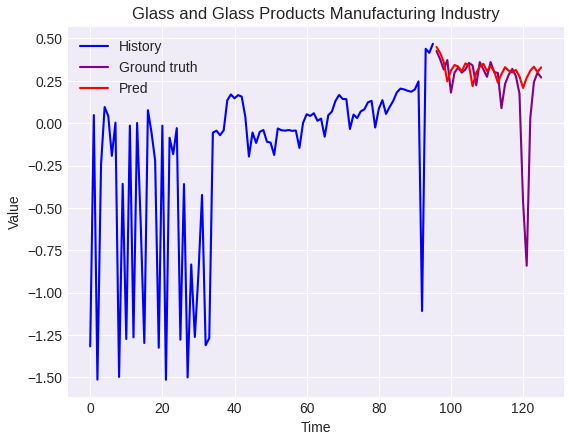}
    \label{fig:subfig9}
  \end{subfigure}
  30-day forecasting of our model.
  \vspace{1em}
  
  \begin{subfigure}[b]{0.32\textwidth}
    \includegraphics[width=\textwidth]{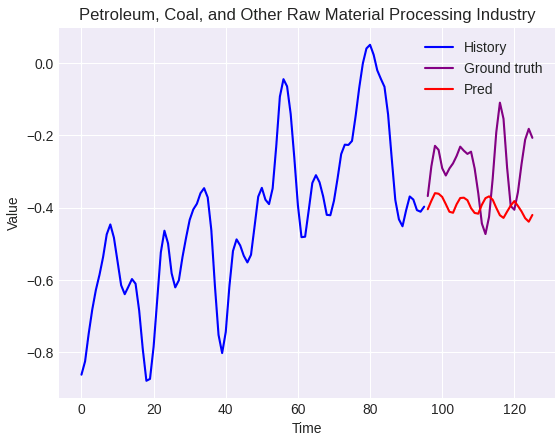}
    \label{fig:subfig10}
  \end{subfigure}
  \hfill
  \begin{subfigure}[b]{0.32\textwidth}
    \includegraphics[width=\textwidth]{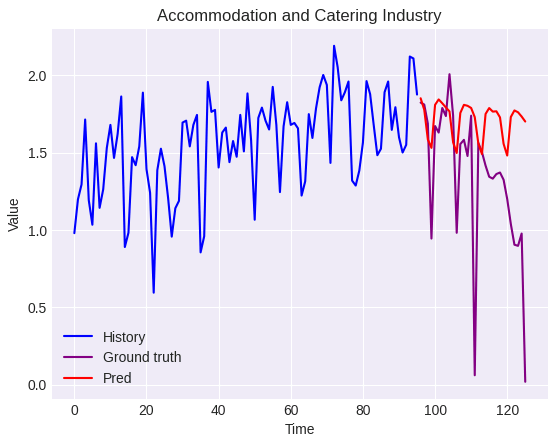}
    \label{fig:subfig11}
  \end{subfigure}
  \hfill
  \begin{subfigure}[b]{0.32\textwidth}
    \includegraphics[width=\textwidth]{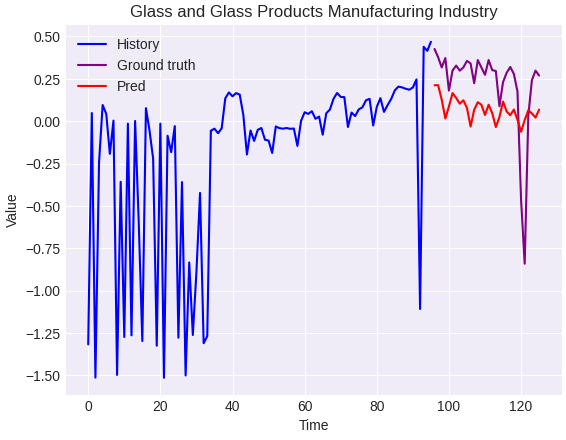}
    \label{fig:subfig12}
  \end{subfigure}
  30-day forecasting of patchtst.
  \caption{Visual comparison of our model and patchtst forecasts.}
  \label{fig:visual}
\end{figure*}

\subsubsection{Visualization analysis}
\par
In Figure \ref{fig:visual}, we show the visualized forecast results of our model and PatchTST for different industries at different forecast horizons. The users from three different industries presented clearly show the significant differences in gas usage characteristics between the various industries. For example, the petroleum, coal, and other raw material processing industry exhibits more continuous and regular changes in gas use; the accommodation and catering industry experiences frequent and unstable fluctuations, making accurate forecasts more challenging; the glass and glass products manufacturing industry sometimes shows frequent changes with large amplitudes, while at other times, it remains relatively stable and regular. After denormalization, it can be observed that the petroleum, coal, and other raw material processing industry and the glass and glass products manufacturing industry typically have higher gas consumption, whereas the accommodation and catering industry, despite significant fluctuations, generally has lower gas consumption. 
\par
In most cases, our model's forecasts are closer to the actual values, demonstrating superior forecasting performance. This is precisely because our foundation model is trained with a large amount of gas usage data, and industry relevance is taken into account to do the corresponding contrastive learning and false-negative sample removal, so it can better adapt to the differences between different industries and show better forecast results.

\subsubsection{Ablation study}
\begin{figure}
    \centering
    \includegraphics[width=0.45\textwidth]{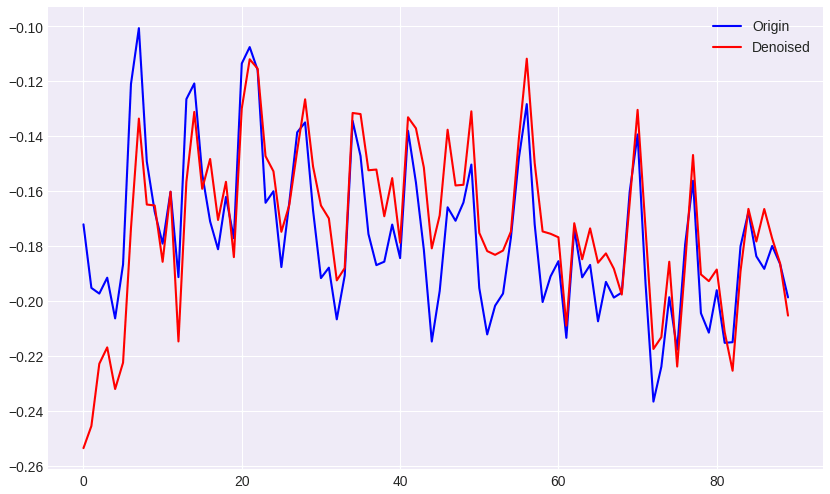}
    \caption{Comparison before and after denoising.}
    \label{fig:denoised}
\end{figure}
For each of the pre-trained components mentioned in this paper, we also performed the corresponding ablation experiments, as shown in Table\ref{tab:ablation}. Due to dataset limitations, we were unable to conduct ablation experiments using the longer historical and forecast horizons employed by PITS\cite{lee2023learning}. Therefore, our ablation experiments were conducted with a historical horizon of 96 and a forecast horizon of 180. In the table, CL represents the contrastive learning component, DN represents the denoising component, and FN represents the false negative sample removal component. The results show that each of our proposed components has a different degree of impact on the model performance, which proves that the design of our individual components is effective. It is worth noting that the performance decline caused by removing CL is similar to that caused by removing FN, which demonstrates the effectiveness of our false negative sample removal design. Without removing these samples, the contrastive learning component would learn incorrect information, leading to minimal improvement.
\par
Figure \ref{fig:denoised} illustrates the comparison of a serie before and after denoising. With the denosing decoder and denoising head, our model filters out the noise from data, allowing the model to make more accurate forecasts, as demonstrated by both our main and ablation experiments.
\begin{table}[pos=htbp]
  \centering
  \caption{Result of ablation experiments. We conducted our ablation experiments using 96 as the historical horizon and 180 as the forecast horizon.}
  \label{tab:ablation}
  \begin{tabular}{c|c|cccc}
    \toprule
    &  & MSE & MAE & SMAPE & MASE \\
    \midrule
    \multirow{4}{*}{Ours} & full  & \textbf{0.6375} & \textbf{0.5571} & \underline{0.2618} & \textbf{1.4408} \\
    & w/o CL & 0.6409 & 0.5579 & 0.2624 & 1.4432 \\
    & w/o DN & 0.6419 & 0.5589 & 0.2623 & 1.443 \\ 
    & w/o FN & 0.6408 & 0.5579 & \textbf{0.2617} & 1.4438 \\
    \bottomrule
  \end{tabular}
\end{table}

\section{Conclusions} \label{Conclusions}
This study introduces a novel foundation model for natural gas demand forecasting, integrating noise filtering and contrastive learning to address the unique challenges posed by industrial and commercial data. Our method effectively mitigates the impact of noise in real-world datasets, enhancing the model's ability to capture meaningful patterns in historical gas usage data. The experimental results, validated on a large-scale, multi-user dataset provided by ENN Group, demonstrate the superior performance of our foundation model across short-term, medium-term, and long-term forecasting scenarios, consistently outperforming existing baseline models.

The proposed foundation model offers significant potential for real-world industrial applications, particularly in the energy sector, where accurate demand forecasting is critical. By improving prediction accuracy and robustness, our approach can contribute to more efficient energy management and resource allocation. Future research could explore extending this foundation model to other forms of energy demand forecasting, such as electricity or water, and enhancing its generalizability across different regions. Additionally, integrating external factors like weather data and economic indicators, as well as adapting the model for real-time forecasting environments, presents promising avenues for further development.

\bibliographystyle{cas-model2-names}

\bibliography{Main}

\begin{thebibliography}{57}
\expandafter\ifx\csname natexlab\endcsname\relax\def\natexlab#1{#1}\fi
\providecommand{\url}[1]{\texttt{#1}}
\providecommand{\href}[2]{#2}
\providecommand{\path}[1]{#1}
\providecommand{\DOIprefix}{doi:}
\providecommand{\ArXivprefix}{arXiv:}
\providecommand{\URLprefix}{URL: }
\providecommand{\Pubmedprefix}{pmid:}
\providecommand{\doi}[1]{\href{http://dx.doi.org/#1}{\path{#1}}}
\providecommand{\Pubmed}[1]{\href{pmid:#1}{\path{#1}}}
\providecommand{\bibinfo}[2]{#2}
\ifx\xfnm\relax \def\xfnm[#1]{\unskip,\space#1}\fi
\bibitem[{Akpinar and Yumusak(2013)}]{akpinar2013forecasting}
\bibinfo{author}{Akpinar, M.}, \bibinfo{author}{Yumusak, N.}, \bibinfo{year}{2013}.
\newblock \bibinfo{title}{Forecasting household natural gas consumption with arima model: A case study of removing cycle}, in: \bibinfo{booktitle}{2013 7th international conference on application of information and communication technologies}, \bibinfo{organization}{IEEE}. pp. \bibinfo{pages}{1--6}.
\bibitem[{Akpinar and Yumusak(2017)}]{akpinar2017day}
\bibinfo{author}{Akpinar, M.}, \bibinfo{author}{Yumusak, N.}, \bibinfo{year}{2017}.
\newblock \bibinfo{title}{Day-ahead natural gas forecasting using nonseasonal exponential smoothing methods}, in: \bibinfo{booktitle}{2017 IEEE International Conference on Environment and Electrical Engineering and 2017 IEEE Industrial and Commercial Power Systems Europe (EEEIC/I\&CPS Europe)}, \bibinfo{organization}{IEEE}. pp. \bibinfo{pages}{1--4}.
\bibitem[{Aydin(2015)}]{aydin2015forecasting}
\bibinfo{author}{Aydin, G.}, \bibinfo{year}{2015}.
\newblock \bibinfo{title}{Forecasting natural gas production using various regression models}.
\newblock \bibinfo{journal}{Petroleum Science and Technology} \bibinfo{volume}{33}, \bibinfo{pages}{1486--1492}.
\bibitem[{Azur et~al.(2011)Azur, Stuart, Frangakis and Leaf}]{azur2011multiple}
\bibinfo{author}{Azur, M.J.}, \bibinfo{author}{Stuart, E.A.}, \bibinfo{author}{Frangakis, C.}, \bibinfo{author}{Leaf, P.J.}, \bibinfo{year}{2011}.
\newblock \bibinfo{title}{Multiple imputation by chained equations: what is it and how does it work?}
\newblock \bibinfo{journal}{International journal of methods in psychiatric research} \bibinfo{volume}{20}, \bibinfo{pages}{40--49}.
\bibitem[{Bai and Li(2016)}]{bai2016daily}
\bibinfo{author}{Bai, Y.}, \bibinfo{author}{Li, C.}, \bibinfo{year}{2016}.
\newblock \bibinfo{title}{Daily natural gas consumption forecasting based on a structure-calibrated support vector regression approach}.
\newblock \bibinfo{journal}{Energy and Buildings} \bibinfo{volume}{127}, \bibinfo{pages}{571--579}.
\bibitem[{Cao et~al.(2018)Cao, Wang, Li, Zhou, Li and Li}]{cao2018brits}
\bibinfo{author}{Cao, W.}, \bibinfo{author}{Wang, D.}, \bibinfo{author}{Li, J.}, \bibinfo{author}{Zhou, H.}, \bibinfo{author}{Li, L.}, \bibinfo{author}{Li, Y.}, \bibinfo{year}{2018}.
\newblock \bibinfo{title}{Brits: Bidirectional recurrent imputation for time series}.
\newblock \bibinfo{journal}{Advances in neural information processing systems} \bibinfo{volume}{31}.
\bibitem[{Chen et~al.(2021)Chen, Peng, Fu and Ling}]{chen2021autoformer}
\bibinfo{author}{Chen, M.}, \bibinfo{author}{Peng, H.}, \bibinfo{author}{Fu, J.}, \bibinfo{author}{Ling, H.}, \bibinfo{year}{2021}.
\newblock \bibinfo{title}{Autoformer: Searching transformers for visual recognition}, in: \bibinfo{booktitle}{Proceedings of the IEEE/CVF international conference on computer vision}, pp. \bibinfo{pages}{12270--12280}.
\bibitem[{Chen et~al.(2024a)Chen, Zhang, Cheng, Shu, Wang, Wen, Yang and Guo}]{chen2024pathformer}
\bibinfo{author}{Chen, P.}, \bibinfo{author}{Zhang, Y.}, \bibinfo{author}{Cheng, Y.}, \bibinfo{author}{Shu, Y.}, \bibinfo{author}{Wang, Y.}, \bibinfo{author}{Wen, Q.}, \bibinfo{author}{Yang, B.}, \bibinfo{author}{Guo, C.}, \bibinfo{year}{2024}a.
\newblock \bibinfo{title}{Pathformer: Multi-scale transformers with adaptive pathways for time series forecasting}.
\newblock \bibinfo{journal}{arXiv preprint arXiv:2402.05956} .
\bibitem[{Chen et~al.(2024b)Chen, Ding, Wang, Xin, Mo, Wang, Han, Luo, Zeng and Wang}]{chen2024context}
\bibinfo{author}{Chen, X.}, \bibinfo{author}{Ding, M.}, \bibinfo{author}{Wang, X.}, \bibinfo{author}{Xin, Y.}, \bibinfo{author}{Mo, S.}, \bibinfo{author}{Wang, Y.}, \bibinfo{author}{Han, S.}, \bibinfo{author}{Luo, P.}, \bibinfo{author}{Zeng, G.}, \bibinfo{author}{Wang, J.}, \bibinfo{year}{2024}b.
\newblock \bibinfo{title}{Context autoencoder for self-supervised representation learning}.
\newblock \bibinfo{journal}{International Journal of Computer Vision} \bibinfo{volume}{132}, \bibinfo{pages}{208--223}.
\bibitem[{Chen et~al.(2018)Chen, Chua and Koch}]{chen2018forecasting}
\bibinfo{author}{Chen, Y.}, \bibinfo{author}{Chua, W.S.}, \bibinfo{author}{Koch, T.}, \bibinfo{year}{2018}.
\newblock \bibinfo{title}{Forecasting day-ahead high-resolution natural-gas demand and supply in germany}.
\newblock \bibinfo{journal}{Applied energy} \bibinfo{volume}{228}, \bibinfo{pages}{1091--1110}.
\bibitem[{Darlow et~al.(2024)Darlow, Deng, Hassan, Asenov, Singh, Joosen, Barker and Storkey}]{darlow2024dam}
\bibinfo{author}{Darlow, L.}, \bibinfo{author}{Deng, Q.}, \bibinfo{author}{Hassan, A.}, \bibinfo{author}{Asenov, M.}, \bibinfo{author}{Singh, R.}, \bibinfo{author}{Joosen, A.}, \bibinfo{author}{Barker, A.}, \bibinfo{author}{Storkey, A.}, \bibinfo{year}{2024}.
\newblock \bibinfo{title}{Dam: Towards a foundation model for time series forecasting}.
\newblock \bibinfo{journal}{arXiv preprint arXiv:2407.17880} .
\bibitem[{Devlin et~al.(2019)Devlin, Chang, Lee and Toutanova}]{devlin2019bert}
\bibinfo{author}{Devlin, J.}, \bibinfo{author}{Chang, M.W.}, \bibinfo{author}{Lee, K.}, \bibinfo{author}{Toutanova, K.}, \bibinfo{year}{2019}.
\newblock \bibinfo{title}{Bert: Pre-training of deep bidirectional transformers for language understanding}, in: \bibinfo{booktitle}{Proceedings of the 2019 Conference of the North American Chapter of the Association for Computational Linguistics: Human Language Technologies, Volume 1 (Long and Short Papers)}, pp. \bibinfo{pages}{4171--4186}.
\bibitem[{Ding et~al.(2023)Ding, Zhao and Jin}]{ding2023forecasting}
\bibinfo{author}{Ding, J.}, \bibinfo{author}{Zhao, Y.}, \bibinfo{author}{Jin, J.}, \bibinfo{year}{2023}.
\newblock \bibinfo{title}{Forecasting natural gas consumption with multiple seasonal patterns}.
\newblock \bibinfo{journal}{Applied Energy} \bibinfo{volume}{337}, \bibinfo{pages}{120911}.
\bibitem[{Dong et~al.(2024)Dong, Wu, Zhang, Zhang, Wang and Long}]{dong2024simmtm}
\bibinfo{author}{Dong, J.}, \bibinfo{author}{Wu, H.}, \bibinfo{author}{Zhang, H.}, \bibinfo{author}{Zhang, L.}, \bibinfo{author}{Wang, J.}, \bibinfo{author}{Long, M.}, \bibinfo{year}{2024}.
\newblock \bibinfo{title}{Simmtm: A simple pre-training framework for masked time-series modeling}.
\newblock \bibinfo{journal}{Advances in Neural Information Processing Systems} \bibinfo{volume}{36}.
\bibitem[{Du et~al.(2023)Du, C{\^o}t{\'e} and Liu}]{du2023saits}
\bibinfo{author}{Du, W.}, \bibinfo{author}{C{\^o}t{\'e}, D.}, \bibinfo{author}{Liu, Y.}, \bibinfo{year}{2023}.
\newblock \bibinfo{title}{Saits: Self-attention-based imputation for time series}.
\newblock \bibinfo{journal}{Expert Systems with Applications} \bibinfo{volume}{219}, \bibinfo{pages}{119619}.
\bibitem[{Duan et~al.(2024)Duan, Zhang, Wu, Cui and Duan}]{GJJJ202403003}
\bibinfo{author}{Duan, Z.}, \bibinfo{author}{Zhang, X.}, \bibinfo{author}{Wu, M.}, \bibinfo{author}{Cui, Z.}, \bibinfo{author}{Duan, T.}, \bibinfo{year}{2024}.
\newblock \bibinfo{title}{China’s natural gas markets review in 2023 and outlook for 2024}.
\newblock \bibinfo{journal}{International Petroleum Economics} \bibinfo{volume}{32}, \bibinfo{pages}{19--28}.
\bibitem[{Ediger and Akar(2007)}]{ediger2007arima}
\bibinfo{author}{Ediger, V.{\c{S}}.}, \bibinfo{author}{Akar, S.}, \bibinfo{year}{2007}.
\newblock \bibinfo{title}{Arima forecasting of primary energy demand by fuel in turkey}.
\newblock \bibinfo{journal}{Energy policy} \bibinfo{volume}{35}, \bibinfo{pages}{1701--1708}.
\bibitem[{Erdogdu(2010)}]{erdogdu2010natural}
\bibinfo{author}{Erdogdu, E.}, \bibinfo{year}{2010}.
\newblock \bibinfo{title}{Natural gas demand in turkey}.
\newblock \bibinfo{journal}{Applied Energy} \bibinfo{volume}{87}, \bibinfo{pages}{211--219}.
\bibitem[{Gorucu(2004)}]{gorucu2004artificial}
\bibinfo{author}{Gorucu, F.}, \bibinfo{year}{2004}.
\newblock \bibinfo{title}{Artificial neural network modeling for forecasting gas consumption}.
\newblock \bibinfo{journal}{Energy Sources} \bibinfo{volume}{26}, \bibinfo{pages}{299--307}.
\bibitem[{Graham(2009)}]{graham2009missing}
\bibinfo{author}{Graham, J.W.}, \bibinfo{year}{2009}.
\newblock \bibinfo{title}{Missing data analysis: Making it work in the real world}.
\newblock \bibinfo{journal}{Annual review of psychology} \bibinfo{volume}{60}, \bibinfo{pages}{549--576}.
\bibitem[{He et~al.(2022)He, Chen, Xie, Li, Doll{\'a}r and Girshick}]{he2022masked}
\bibinfo{author}{He, K.}, \bibinfo{author}{Chen, X.}, \bibinfo{author}{Xie, S.}, \bibinfo{author}{Li, Y.}, \bibinfo{author}{Doll{\'a}r, P.}, \bibinfo{author}{Girshick, R.}, \bibinfo{year}{2022}.
\newblock \bibinfo{title}{Masked autoencoders are scalable vision learners}, in: \bibinfo{booktitle}{Proceedings of the IEEE/CVF conference on computer vision and pattern recognition}, pp. \bibinfo{pages}{16000--16009}.
\bibitem[{Ho et~al.(2020)Ho, Jain and Abbeel}]{ho2020denoising}
\bibinfo{author}{Ho, J.}, \bibinfo{author}{Jain, A.}, \bibinfo{author}{Abbeel, P.}, \bibinfo{year}{2020}.
\newblock \bibinfo{title}{Denoising diffusion probabilistic models}.
\newblock \bibinfo{journal}{Advances in neural information processing systems} \bibinfo{volume}{33}, \bibinfo{pages}{6840--6851}.
\bibitem[{Hou and Nguyen(2018)}]{hou2018understanding}
\bibinfo{author}{Hou, C.}, \bibinfo{author}{Nguyen, B.H.}, \bibinfo{year}{2018}.
\newblock \bibinfo{title}{Understanding the us natural gas market: A markov switching var approach}.
\newblock \bibinfo{journal}{Energy Economics} \bibinfo{volume}{75}, \bibinfo{pages}{42--53}.
\bibitem[{Hussain et~al.(2022)Hussain, Memon, Murshed, Alam, Mehmood, Alam, Rahman and Hayat}]{hussain2022time}
\bibinfo{author}{Hussain, A.}, \bibinfo{author}{Memon, J.A.}, \bibinfo{author}{Murshed, M.}, \bibinfo{author}{Alam, M.S.}, \bibinfo{author}{Mehmood, U.}, \bibinfo{author}{Alam, M.N.}, \bibinfo{author}{Rahman, M.}, \bibinfo{author}{Hayat, U.}, \bibinfo{year}{2022}.
\newblock \bibinfo{title}{A time series forecasting analysis of overall and sector-based natural gas demand: a developing south asian economy case}.
\newblock \bibinfo{journal}{Environmental Science and Pollution Research} \bibinfo{volume}{29}, \bibinfo{pages}{71676--71687}.
\bibitem[{Khotanzad et~al.(2000)Khotanzad, Elragal and Lu}]{khotanzad2000combination}
\bibinfo{author}{Khotanzad, A.}, \bibinfo{author}{Elragal, H.}, \bibinfo{author}{Lu, T.L.}, \bibinfo{year}{2000}.
\newblock \bibinfo{title}{Combination of artificial neural-network forecasters for prediction of natural gas consumption}.
\newblock \bibinfo{journal}{IEEE transactions on neural networks} \bibinfo{volume}{11}, \bibinfo{pages}{464--473}.
\bibitem[{Kim et~al.(2021)Kim, Kim, Tae, Park, Choi and Choo}]{kim2021reversible}
\bibinfo{author}{Kim, T.}, \bibinfo{author}{Kim, J.}, \bibinfo{author}{Tae, Y.}, \bibinfo{author}{Park, C.}, \bibinfo{author}{Choi, J.H.}, \bibinfo{author}{Choo, J.}, \bibinfo{year}{2021}.
\newblock \bibinfo{title}{Reversible instance normalization for accurate time-series forecasting against distribution shift}, in: \bibinfo{booktitle}{International Conference on Learning Representations}.
\bibitem[{Laib et~al.(2019)Laib, Khadir and Mihaylova}]{laib2019toward}
\bibinfo{author}{Laib, O.}, \bibinfo{author}{Khadir, M.T.}, \bibinfo{author}{Mihaylova, L.}, \bibinfo{year}{2019}.
\newblock \bibinfo{title}{Toward efficient energy systems based on natural gas consumption prediction with lstm recurrent neural networks}.
\newblock \bibinfo{journal}{Energy} \bibinfo{volume}{177}, \bibinfo{pages}{530--542}.
\bibitem[{Lee et~al.(2023)Lee, Park and Lee}]{lee2023learning}
\bibinfo{author}{Lee, S.}, \bibinfo{author}{Park, T.}, \bibinfo{author}{Lee, K.}, \bibinfo{year}{2023}.
\newblock \bibinfo{title}{Learning to embed time series patches independently}.
\newblock \bibinfo{journal}{arXiv preprint arXiv:2312.16427} .
\bibitem[{Lin et~al.(2024)Lin, Xie and Zhang}]{lin2024compound}
\bibinfo{author}{Lin, Z.}, \bibinfo{author}{Xie, L.}, \bibinfo{author}{Zhang, S.}, \bibinfo{year}{2024}.
\newblock \bibinfo{title}{A compound framework for short-term gas load forecasting combining time-enhanced perception transformer and two-stage feature extraction}.
\newblock \bibinfo{journal}{Energy} \bibinfo{volume}{298}, \bibinfo{pages}{131365}.
\bibitem[{Liu et~al.(2021a)Liu, Yu, Liao, Li, Lin, Liu and Dustdar}]{liu2021pyraformer}
\bibinfo{author}{Liu, S.}, \bibinfo{author}{Yu, H.}, \bibinfo{author}{Liao, C.}, \bibinfo{author}{Li, J.}, \bibinfo{author}{Lin, W.}, \bibinfo{author}{Liu, A.X.}, \bibinfo{author}{Dustdar, S.}, \bibinfo{year}{2021}a.
\newblock \bibinfo{title}{Pyraformer: Low-complexity pyramidal attention for long-range time series modeling and forecasting}, in: \bibinfo{booktitle}{International conference on learning representations}.
\bibitem[{Liu et~al.(2021b)Liu, Liang, Zheng, Hooi and Zimmermann}]{liu2021spatio}
\bibinfo{author}{Liu, X.}, \bibinfo{author}{Liang, Y.}, \bibinfo{author}{Zheng, Y.}, \bibinfo{author}{Hooi, B.}, \bibinfo{author}{Zimmermann, R.}, \bibinfo{year}{2021}b.
\newblock \bibinfo{title}{Spatio-temporal graph contrastive learning}.
\newblock \bibinfo{journal}{arXiv preprint arXiv:2108.11873} .
\bibitem[{Liu et~al.(2023)Liu, Hu, Zhang, Wu, Wang, Ma and Long}]{liu2023itransformer}
\bibinfo{author}{Liu, Y.}, \bibinfo{author}{Hu, T.}, \bibinfo{author}{Zhang, H.}, \bibinfo{author}{Wu, H.}, \bibinfo{author}{Wang, S.}, \bibinfo{author}{Ma, L.}, \bibinfo{author}{Long, M.}, \bibinfo{year}{2023}.
\newblock \bibinfo{title}{itransformer: Inverted transformers are effective for time series forecasting}.
\newblock \bibinfo{journal}{arXiv preprint arXiv:2310.06625} .
\bibitem[{Liu et~al.(2022)Liu, Wu, Wang and Long}]{liu2022non}
\bibinfo{author}{Liu, Y.}, \bibinfo{author}{Wu, H.}, \bibinfo{author}{Wang, J.}, \bibinfo{author}{Long, M.}, \bibinfo{year}{2022}.
\newblock \bibinfo{title}{Non-stationary transformers: Exploring the stationarity in time series forecasting}.
\newblock \bibinfo{journal}{Advances in Neural Information Processing Systems} \bibinfo{volume}{35}, \bibinfo{pages}{9881--9893}.
\bibitem[{Merkel et~al.(2018)Merkel, Povinelli and Brown}]{merkel2018short}
\bibinfo{author}{Merkel, G.D.}, \bibinfo{author}{Povinelli, R.J.}, \bibinfo{author}{Brown, R.H.}, \bibinfo{year}{2018}.
\newblock \bibinfo{title}{Short-term load forecasting of natural gas with deep neural network regression}.
\newblock \bibinfo{journal}{Energies} \bibinfo{volume}{11}, \bibinfo{pages}{2008}.
\bibitem[{Nguyen et~al.(2023)Nguyen, Brandstetter, Kapoor, Gupta and Grover}]{nguyen2023climax}
\bibinfo{author}{Nguyen, T.}, \bibinfo{author}{Brandstetter, J.}, \bibinfo{author}{Kapoor, A.}, \bibinfo{author}{Gupta, J.K.}, \bibinfo{author}{Grover, A.}, \bibinfo{year}{2023}.
\newblock \bibinfo{title}{Climax: A foundation model for weather and climate}.
\newblock \bibinfo{journal}{arXiv preprint arXiv:2301.10343} .
\bibitem[{Nie et~al.(2022)Nie, Nguyen, Sinthong and Kalagnanam}]{nie2022time}
\bibinfo{author}{Nie, Y.}, \bibinfo{author}{Nguyen, N.H.}, \bibinfo{author}{Sinthong, P.}, \bibinfo{author}{Kalagnanam, J.}, \bibinfo{year}{2022}.
\newblock \bibinfo{title}{A time series is worth 64 words: Long-term forecasting with transformers}.
\newblock \bibinfo{journal}{arXiv preprint arXiv:2211.14730} .
\bibitem[{Niu et~al.(2024)Niu, Xu, Liu, Wen and Yuan}]{niucontrastive}
\bibinfo{author}{Niu, Y.}, \bibinfo{author}{Xu, H.}, \bibinfo{author}{Liu, C.}, \bibinfo{author}{Wen, Y.}, \bibinfo{author}{Yuan, X.}, \bibinfo{year}{2024}.
\newblock \bibinfo{title}{Contrastive representation learning for self-supervised taxonomy completion}, in: \bibinfo{booktitle}{Proceedings of the Thirty-Third International Joint Conference on Artificial Intelligence, {IJCAI-24}}, pp. \bibinfo{pages}{6442--6450}.
\newblock \URLprefix \url{https://doi.org/10.24963/ijcai.2024/712}, \DOIprefix\doi{10.24963/ijcai.2024/712}.
\bibitem[{Petkovic et~al.(2022)Petkovic, Koch and Zittel}]{petkovic2022deep}
\bibinfo{author}{Petkovic, M.}, \bibinfo{author}{Koch, T.}, \bibinfo{author}{Zittel, J.}, \bibinfo{year}{2022}.
\newblock \bibinfo{title}{Deep learning for spatio-temporal supply and demand forecasting in natural gas transmission networks}.
\newblock \bibinfo{journal}{Energy Science \& Engineering} \bibinfo{volume}{10}, \bibinfo{pages}{1812--1825}.
\bibitem[{Pu et~al.(2024)Pu, Zhu, Yang, L{\"u} and Yang}]{pu2024novel}
\bibinfo{author}{Pu, Y.}, \bibinfo{author}{Zhu, C.}, \bibinfo{author}{Yang, K.}, \bibinfo{author}{L{\"u}, Z.}, \bibinfo{author}{Yang, Q.}, \bibinfo{year}{2024}.
\newblock \bibinfo{title}{A novel multiscale transformer network framework for natural gas consumption forecasting}.
\newblock \bibinfo{journal}{IEEE Transactions on Industrial Informatics} .
\bibitem[{Sen et~al.(2019)Sen, G{\"u}nay and Tun{\c{c}}}]{sen2019forecasting}
\bibinfo{author}{Sen, D.}, \bibinfo{author}{G{\"u}nay, M.E.}, \bibinfo{author}{Tun{\c{c}}, K.M.}, \bibinfo{year}{2019}.
\newblock \bibinfo{title}{Forecasting annual natural gas consumption using socio-economic indicators for making future policies}.
\newblock \bibinfo{journal}{Energy} \bibinfo{volume}{173}, \bibinfo{pages}{1106--1118}.
\bibitem[{Song et~al.(2022)Song, Zhang, Jiang, Ma, Zhang, Xue, Shen and Wu}]{song2022estimate}
\bibinfo{author}{Song, J.}, \bibinfo{author}{Zhang, L.}, \bibinfo{author}{Jiang, Q.}, \bibinfo{author}{Ma, Y.}, \bibinfo{author}{Zhang, X.}, \bibinfo{author}{Xue, G.}, \bibinfo{author}{Shen, X.}, \bibinfo{author}{Wu, X.}, \bibinfo{year}{2022}.
\newblock \bibinfo{title}{Estimate the daily consumption of natural gas in district heating system based on a hybrid seasonal decomposition and temporal convolutional network model}.
\newblock \bibinfo{journal}{Applied Energy} \bibinfo{volume}{309}, \bibinfo{pages}{118444}.
\bibitem[{Su et~al.(2019)Su, Zio, Zhang, Chi, Li and Zhang}]{su2019systematic}
\bibinfo{author}{Su, H.}, \bibinfo{author}{Zio, E.}, \bibinfo{author}{Zhang, J.}, \bibinfo{author}{Chi, L.}, \bibinfo{author}{Li, X.}, \bibinfo{author}{Zhang, Z.}, \bibinfo{year}{2019}.
\newblock \bibinfo{title}{A systematic data-driven demand side management method for smart natural gas supply systems}.
\newblock \bibinfo{journal}{Energy Conversion and Management} \bibinfo{volume}{185}, \bibinfo{pages}{368--383}.
\bibitem[{Tian et~al.(2024)Tian, Shao, Bian, Zeng, Li and Zhao}]{tian2024application}
\bibinfo{author}{Tian, N.}, \bibinfo{author}{Shao, B.}, \bibinfo{author}{Bian, G.}, \bibinfo{author}{Zeng, H.}, \bibinfo{author}{Li, X.}, \bibinfo{author}{Zhao, W.}, \bibinfo{year}{2024}.
\newblock \bibinfo{title}{Application of forecasting strategies and techniques to natural gas consumption: A comprehensive review and comparative study}.
\newblock \bibinfo{journal}{Engineering Applications of Artificial Intelligence} \bibinfo{volume}{129}, \bibinfo{pages}{107644}.
\bibitem[{Touvron et~al.(2023)Touvron, Martin, Stone, Albert, Almahairi, Babaei, Bashlykov, Batra, Bhargava, Bhosale et~al.}]{touvron2023llama}
\bibinfo{author}{Touvron, H.}, \bibinfo{author}{Martin, L.}, \bibinfo{author}{Stone, K.}, \bibinfo{author}{Albert, P.}, \bibinfo{author}{Almahairi, A.}, \bibinfo{author}{Babaei, Y.}, \bibinfo{author}{Bashlykov, N.}, \bibinfo{author}{Batra, S.}, \bibinfo{author}{Bhargava, P.}, \bibinfo{author}{Bhosale, S.}, et~al., \bibinfo{year}{2023}.
\newblock \bibinfo{title}{Llama 2: Open foundation and fine-tuned chat models}.
\newblock \bibinfo{journal}{arXiv preprint arXiv:2307.09288} .
\bibitem[{Tu et~al.(2024)Tu, Zhang, Zhang and Yang}]{tu2024powerpm}
\bibinfo{author}{Tu, S.}, \bibinfo{author}{Zhang, Y.}, \bibinfo{author}{Zhang, J.}, \bibinfo{author}{Yang, Y.}, \bibinfo{year}{2024}.
\newblock \bibinfo{title}{Powerpm: Foundation model for power systems}.
\newblock \bibinfo{journal}{arXiv preprint arXiv:2408.04057} .
\bibitem[{Wang et~al.(2024)Wang, Zhou, Wen, Gao, Ding and Jin}]{wang2024card}
\bibinfo{author}{Wang, X.}, \bibinfo{author}{Zhou, T.}, \bibinfo{author}{Wen, Q.}, \bibinfo{author}{Gao, J.}, \bibinfo{author}{Ding, B.}, \bibinfo{author}{Jin, R.}, \bibinfo{year}{2024}.
\newblock \bibinfo{title}{Card: Channel aligned robust blend transformer for time series forecasting}, in: \bibinfo{booktitle}{The Twelfth International Conference on Learning Representations}.
\bibitem[{White and Carlin(2010)}]{white2010bias}
\bibinfo{author}{White, I.R.}, \bibinfo{author}{Carlin, J.B.}, \bibinfo{year}{2010}.
\newblock \bibinfo{title}{Bias and efficiency of multiple imputation compared with complete-case analysis for missing covariate values}.
\newblock \bibinfo{journal}{Statistics in medicine} \bibinfo{volume}{29}, \bibinfo{pages}{2920--2931}.
\bibitem[{Wu et~al.(2022)Wu, Hu, Liu, Zhou, Wang and Long}]{wu2022timesnet}
\bibinfo{author}{Wu, H.}, \bibinfo{author}{Hu, T.}, \bibinfo{author}{Liu, Y.}, \bibinfo{author}{Zhou, H.}, \bibinfo{author}{Wang, J.}, \bibinfo{author}{Long, M.}, \bibinfo{year}{2022}.
\newblock \bibinfo{title}{Timesnet: Temporal 2d-variation modeling for general time series analysis}.
\newblock \bibinfo{journal}{arXiv preprint arXiv:2210.02186} .
\bibitem[{Wu et~al.(2019)Wu, Wang and Wang}]{wu2019self}
\bibinfo{author}{Wu, J.}, \bibinfo{author}{Wang, X.}, \bibinfo{author}{Wang, W.Y.}, \bibinfo{year}{2019}.
\newblock \bibinfo{title}{Self-supervised dialogue learning}, in: \bibinfo{booktitle}{Proceedings of the 57th Annual Meeting of the Association for Computational Linguistics}, pp. \bibinfo{pages}{3857--3867}.
\bibitem[{Xue et~al.(2019)Xue, Song, Kong, Pan, Qi and Li}]{xue2019prediction}
\bibinfo{author}{Xue, G.}, \bibinfo{author}{Song, J.}, \bibinfo{author}{Kong, X.}, \bibinfo{author}{Pan, Y.}, \bibinfo{author}{Qi, C.}, \bibinfo{author}{Li, H.}, \bibinfo{year}{2019}.
\newblock \bibinfo{title}{Prediction of natural gas consumption for city-level dhs based on attention gru: A case study for a northern chinese city}.
\newblock \bibinfo{journal}{IEEE Access} \bibinfo{volume}{7}, \bibinfo{pages}{130685--130699}.
\bibitem[{Yang et~al.(2022)Yang, Zhang and Cui}]{yang2022timeclr}
\bibinfo{author}{Yang, X.}, \bibinfo{author}{Zhang, Z.}, \bibinfo{author}{Cui, R.}, \bibinfo{year}{2022}.
\newblock \bibinfo{title}{Timeclr: A self-supervised contrastive learning framework for univariate time series representation}.
\newblock \bibinfo{journal}{Knowledge-Based Systems} \bibinfo{volume}{245}, \bibinfo{pages}{108606}.
\bibitem[{Yu et~al.(2016)Yu, Rao and Dhillon}]{yu2016temporal}
\bibinfo{author}{Yu, H.F.}, \bibinfo{author}{Rao, N.}, \bibinfo{author}{Dhillon, I.S.}, \bibinfo{year}{2016}.
\newblock \bibinfo{title}{Temporal regularized matrix factorization for high-dimensional time series prediction}.
\newblock \bibinfo{journal}{Advances in neural information processing systems} \bibinfo{volume}{29}.
\bibitem[{Zeng et~al.(2023)Zeng, Chen, Zhang and Xu}]{zeng2023transformers}
\bibinfo{author}{Zeng, A.}, \bibinfo{author}{Chen, M.}, \bibinfo{author}{Zhang, L.}, \bibinfo{author}{Xu, Q.}, \bibinfo{year}{2023}.
\newblock \bibinfo{title}{Are transformers effective for time series forecasting?}, in: \bibinfo{booktitle}{Proceedings of the AAAI conference on artificial intelligence}, pp. \bibinfo{pages}{11121--11128}.
\bibitem[{Zhang and Yan(2022)}]{zhang2022crossformer}
\bibinfo{author}{Zhang, Y.}, \bibinfo{author}{Yan, J.}, \bibinfo{year}{2022}.
\newblock \bibinfo{title}{Crossformer: Transformer utilizing cross-dimension dependency for multivariate time series forecasting}, in: \bibinfo{booktitle}{The eleventh international conference on learning representations}.
\bibitem[{Zhou et~al.(2021)Zhou, Zhang, Peng, Zhang, Li, Xiong and Zhang}]{zhou2021informer}
\bibinfo{author}{Zhou, H.}, \bibinfo{author}{Zhang, S.}, \bibinfo{author}{Peng, J.}, \bibinfo{author}{Zhang, S.}, \bibinfo{author}{Li, J.}, \bibinfo{author}{Xiong, H.}, \bibinfo{author}{Zhang, W.}, \bibinfo{year}{2021}.
\newblock \bibinfo{title}{Informer: Beyond efficient transformer for long sequence time-series forecasting}, in: \bibinfo{booktitle}{Proceedings of the AAAI conference on artificial intelligence}, pp. \bibinfo{pages}{11106--11115}.
\bibitem[{Zhou et~al.(2022)Zhou, Ma, Wen, Wang, Sun and Jin}]{zhou2022fedformer}
\bibinfo{author}{Zhou, T.}, \bibinfo{author}{Ma, Z.}, \bibinfo{author}{Wen, Q.}, \bibinfo{author}{Wang, X.}, \bibinfo{author}{Sun, L.}, \bibinfo{author}{Jin, R.}, \bibinfo{year}{2022}.
\newblock \bibinfo{title}{Fedformer: Frequency enhanced decomposed transformer for long-term series forecasting}, in: \bibinfo{booktitle}{International conference on machine learning}, \bibinfo{organization}{PMLR}. pp. \bibinfo{pages}{27268--27286}.
\bibitem[{Zhu et~al.(2015)Zhu, Li, Wu and Jiang}]{zhu2015short}
\bibinfo{author}{Zhu, L.}, \bibinfo{author}{Li, M.}, \bibinfo{author}{Wu, Q.}, \bibinfo{author}{Jiang, L.}, \bibinfo{year}{2015}.
\newblock \bibinfo{title}{Short-term natural gas demand prediction based on support vector regression with false neighbours filtered}.
\newblock \bibinfo{journal}{Energy} \bibinfo{volume}{80}, \bibinfo{pages}{428--436}.

\end{thebibliography}



\end{document}